\documentclass[11pt, a4paper, twocolumn]{article}

\usepackage[margin=0.8in]{geometry}
\usepackage{authblk}
\usepackage[utf8]{inputenc} 
\usepackage[T1]{fontenc}    
\usepackage{hyperref}       
\usepackage{url}            
\usepackage{booktabs}       
\usepackage{amsfonts}       
\usepackage{nicefrac}       
\usepackage{enumitem}
\usepackage{microtype}
\usepackage{graphicx}
\usepackage{subcaption}
\usepackage[square,sort,comma,numbers]{natbib}
\usepackage{float}
\usepackage{amssymb,amsmath,color,bm,algorithm,algorithmic}

\usepackage[linesnumbered,ruled,vlined,algo2e]{algorithm2e}

\SetCommentSty{mycommfont}

\SetKwInput{KwInput}{Input}                
\SetKwInput{KwOutput}{Output}              

\graphicspath{{figs/}}

\usepackage{url}

\usepackage[mathcal]{eucal}

\DeclareMathAlphabet\mathbfcal{OMS}{cmsy}{b}{n}

\renewcommand{\hat}{\widehat}
\newcommand{\BEAS}{\begin{eqnarray*}}
\newcommand{\EEAS}{\end{eqnarray*}}
\newcommand{\BEA}{\begin{eqnarray}}
\newcommand{\EEA}{\end{eqnarray}}
\newcommand{\BEQ}{\begin{equation}}
\newcommand{\EEQ}{\end{equation}}
\newcommand{\BIT}{\begin{itemize}}
\newcommand{\EIT}{\end{itemize}}
\newcommand{\BNUM}{\begin{enumerate}}
\newcommand{\ENUM}{\end{enumerate}}
\newcommand{\BA}{\begin{array}}
\newcommand{\EA}{\end{array}}

\newcommand{\argmin}{\mathop{\rm argmin}}

\newcommand{\BlackBox}{\rule{1.5ex}{1.5ex}}  
\newcommand{\cA}{\mathcal{A}}

\newenvironment{proof}{\par\noindent{\bf Proof\ }}{\hfill\BlackBox\\[2mm]}
\newtheorem{lemma}{Lemma}
\newtheorem{theorem}{Theorem}
\newtheorem{proposition}{Proposition}

\def \E{{\mathbb E}}

\allowdisplaybreaks[4]
\newcommand{\punt}[1]{}

\newtheorem{remark}{Remark}

\def\argmin{\mathop{\rm arg\,min}}

\newcommand{\reals}{\mathbb{R}}

\newcommand{\co}{\mathcal{O}}

\def\argmin{\mathop{\rm arg\,min}}

\newcommand{\bbE}{\mathbb{E}}

\newcommand{\bq}{\begin{equation}}
\newcommand{\ba}{\begin{eqnarray}}
\newcommand{\ea}{\end{eqnarray}}

\def\R{{\reals}}

\newcommand{\remove}[1]{}

\usepackage{cleveref}

\newcommand{\eqdef}{:=}

\title{Non-stationary Online  Regression}

\author[1]{Anant Raj \thanks{raj.anant12@gmail.com}}
\author[2]{Pierre Gaillard \thanks{pierre.gaillard@inria.fr}}
\author[3]{Christophe Saad\thanks{xophesaad@gmail.com}}
\affil[1]{Max-Planck Institute for Intelligent Systems, T\"ubingen}
\affil[2]{Univ. Grenoble Alpes, Inria, CNRS, Grenoble INP, LJK, 38000 Grenoble, France.}
\affil[3]{Carnegie Mellon University}

\begin{document}
\maketitle

\begin{abstract}
 Online forecasting under changing environment has been a problem of increasing importance in many real-world applications. In this paper, we consider the meta-algorithm presented in \citet{zhang2017dynamic} combined with different subroutines. We show that an expected cumulative error of order $\tilde{\mathcal{O}}(n^{1/3} C_n^{2/3})$ can be obtained for non-stationary online linear regression where the total variation of parameter sequence is bounded by $C_n$. Our paper extends the result of online forecasting of one dimensional time-series as proposed in \cite{baby2019online} to general $d$-dimensional non-stationary linear regression. We improve the rate $O(\sqrt{n C_n})$ obtained by \citet{zhang2017dynamic} and \citet{besbes2015non}. We further extend our analysis to non-stationary online kernel regression. Similar to the non-stationary online regression case, we use the meta-procedure of \citet{zhang2017dynamic} combined with Kernel-AWV \citep{jezequel2019efficient} to achieve an expected cumulative controlled by the effective dimension of the RKHS and the total variation of the sequence. To the best of our knowledge, this work is the first extension of non-stationary online regression to non-stationary kernel regression. Lastly, we evaluate our method empirically with several existing benchmarks and also compare it with the theoretical bound obtained in this paper.
\end{abstract}
\vspace{-1mm}
\section{Introduction}\label{sec:intro}
\vspace{-2mm}

We consider online linear regression in a non-stationary environment. More formally, at each round $t=1,\dots,n$, the learner receives an input $x_t \in \R^d$, makes a prediction $\hat y_t \in \R$ and receives a noisy output $y_t = x_t^\top \theta_t + Z_t$ where $\theta_t \in \R^d$ is some unknown parameter and $Z_t$ are i.i.d. sub-Gaussian noise. We are interested in minimizing the expected cumulative error
\begin{equation}
    R_n(\hat {y}_{1:n}, \theta_{1:n}) \eqdef  \sum_{t=1}^n \E \big[ (\hat y_t - x_t^\top \theta_t)^2\big] \,.
    \label{eq:cum_error}
\end{equation}
Of course, without further assumption, the cumulative error is doomed to grow linearly in $n$. Therefore, we assume there is regularity in the signal $\theta_{1:n} = \big(\theta_1,\dots,\theta_n\big) \in \R^{d\times n}$, measured by its total variation
\begin{equation}
    TV(\theta_{1:n}) = \sum_{t=2}^n \big\|\theta_t - \theta_{t-1}\big\|_1 \,.
    \label{eq:total_variation}
\end{equation}
We also assume that there exists $B >0$ such that for all $t\geq 1$, $\|\theta_t\|_1 \leq B$. We emphasize that apart from boundedness in $\ell_1$-norm and in total variation, we do not make any assumption on the sequence $\theta_{1:n}$. The latter is arbitrary and may be chosen by an adversary.  

\paragraph{Related Works} \label{sec:rel_works}
Online prediction of arbitrary time-series has already been well studied by the online learning and optimization communities and we refer to the monographs \cite{cesa2006prediction,hazan2019introduction} and references therein for detailed overviews. A very large part of the existing work only deals with stationary environment, in which the learner's performance is compared with respect to some fixed strategy that does not evolve over time. Thanks to many applications (e.g. web marketing or electricity forecasting), designing strategies that adapt to a changing environment has recently drawn considerable attention. 

Online learning in a non-stationary environment was referred under different names or settings as ``shifting regret'', ``adaptive regret'', ``dynamic regret'', or ``tracking the best predictor'' but most of these notions are strongly related. Some relevant works are \cite{besbes2015non,herbster2001tracking,zinkevich2003online,Cesa-BianchiGaillardLugosiStoltz2012,hazan2009efficient,bousquet2002tracking,jadbabaie2015online,mokhtari2016online,yang2016tracking}.  \cite{herbster2001tracking} first considered  shifting bounds for linear regression using projected mirror descent. \cite{zinkevich2003online} provides dynamic regret guarantees for any convex losses for projected online gradient descent. Most of these work considered however non noisy observations (or gradients), as we consider. 
\cite{besbes2015non} proved matching upper and lower bounds for the dynamic regret with noisy observations. They provide dynamic regret bounds of order $TV(\theta_{1:n})^{1/3} n^{2/3}$ for convex losses and $\sqrt{TV(\theta_{1:n})n}$ for strongly convex losses. The latter was generalized to exp-concave losses by \cite{zhang2017dynamic}.

\paragraph{Contributions} Most of the above works consider the regret, while here we consider the cumulative error~\eqref{eq:cum_error}. In other words, in our case, the performance of the player is only compared with respect to the true underlying sequence $\theta_{1:n}$ which must have low total variation. This assumption allows us to prove stronger guarantees. Indeed, in the one-dimensional setting of online forecasting of a time-series with square loss, \cite{baby2019online} could prove that the optimal rate of order $TV(\theta_{1:n})^{2/3} n^{1/3}$ instead of $\sqrt{TV(\theta_{1:n})n}$ for the cumulative error~\eqref{eq:cum_error}. Their technique is based on change point detection via wavelets and heavily relies on their simple setting (one dimension, no input~$x_t$). 

In this work, we generalize the result of \cite{baby2019online} to online linear regression in dimension $d$ and to reproducing kernel Hilbert spaces (RKHS). We ended up by using the meta-procedures of \cite{hazan2007adaptive} and \cite{zhang2017dynamic} for exp-concave loss functions, combined with well-chosen subroutines. Carrying a careful regret analysis in our setting, we achieve the optimal error of \cite{baby2019online}.

Finally, in Section~\ref{sec:exp}, we corroborate our theoretical results on numerical simulations. 

\vspace{-1mm}
\section{Warm-up: Online Prediction of Non-Stationary Time Series} \label{sec:warmup}
\vspace{-2mm}
In this section, we discuss the relevant background to our work and simple intuition for $1$-dimensional problem. However, before going into the details of our approach, we first discuss the work of \citet{baby2019online} which considers one dimensional non-stationary online linear regression. 
\vspace{-1mm}
\subsection{ARROWS \cite{baby2019online}} \label{sec:arrows}
\vspace{-1mm}
ARROWS considers to solve the problem of online forecasting of sequences of length $n$ whose total-variation (TV) is at most $n$. The observed output is the noise contaminated version of original input sequence $\theta_i$ for $i$ in $[n]$.  ARROWS considers to predict via the moving average of the output in an interval.  If the total-variation within that time interval is small then the moving average in that time interval is reasonably good prediction to minimize the cumulative squared error. For that reason, the algorithm needs to detect intervals which has low total variation. This task of detection is accomplished by constructing a lower bound of TV which acts like a threshold to restart the averaging and hence acts like a non-linearity which can capture the non-linear variation in the sequence. The estimation of the lower bound is based on computing of Haar coefficients  as it smooths the  adjacent regions of a signal and then taking difference between them. A slightly modified version of the soft threshold estimator from
 from \citet{donoho1990minimax}  is considered for oracle estimator.

Overall, the restart strategy based on change point detection using Haar coefficients proposed in this work achieves the optimal error however, the approach is very hard to extend beyond one dimensional regression problem. Another drawback this work has is that ARROWS requires to know the noise level sigma to tun the algorithm even in one dimensional forecasting problem. To know the exact noise level is an unrealistic assumption in real life problems. We address here these two concerns.
\vspace{-1mm}
\subsection{One-Dimensional Intuition} \label{sec:1d}
\vspace{-1mm}
In this section, we consider the simpler case with $d = 1$ and $x_t = 1$ that was already considered by \cite{baby2019online} as  a warm up to understand the intuition behind our algorithm.  Let us now define the formal problem. The problem formulation looks as:
$y_t = \theta_t + Z_t$ for $t = 1,\cdots, n$ and $Z_t$ be independent $\sigma$ sub-Gaussian random variables.   The goal of the problem is to recover $\theta_t$ by minimizing the cumulative error $R_n(\hat y_{1:n}, \theta_{1:n}) = \sum_{t=1}^T \E\big[(\hat y_t - \theta_t)^2\big]$. 
\vspace{-1mm}
\paragraph{Lower-bound and previous results}
In \cite{roy2019online}, the authors first prove that using online gradient descent with fixed restart (as considered by~\cite{besbes2015non}) is sub-optimal in this setting. Their theorem 2 shows a cumulative error for OGD with fixed restart of order $\tilde{\co}(B^2 + TV(\theta_{1:n})^2 + \sigma TV(\theta_{1:n}) \sqrt{n})$, where $B$ is an upper-bound on $\|\theta_1\|_1$. Yet, they also prove the following lower-bound.
\begin{proposition}[{\cite[Proposition 2]{baby2019online}}] \label{prp:lower_bound_oned}
Let $n \geq 2$, $\sigma >0$, and $B, C_n > 0$ such that $\min\{ B,C_n\} > 2\pi \sigma$. Then, there is a universal constant $c$ such that, for any forecaster, there exists a sequence $\theta_1,\dots,\theta_n$ such that $TV(\theta_{1:n}) \leq C_n$ and
\small
 $$ 
   R_n\big(\hat y_{1:n}, \theta_{1:n}\big) \geq c(B^2 + C_n^2  + \sigma^2 \log n + n^{1/3} C_n^{2/3} \sigma^{4/3}).
 $$
 \normalsize
\end{proposition}
Our aim is to address the two major challenges of ARROWS discussed previously (address general $d$-dimensional problems and no need to know the exact noise level $\sigma$) while achieving an optimal error of order $O(n^{1/3}C_n^{2/3})$.  

\vspace{-1mm}
\paragraph{An hypothetical forecaster which achieves optimal error} 
Let $m \geq 1$. We first analyse the approximation error obtained by an hypothetical forecaster that produces moving average with at most $m$ restarts. It first computes a sequence of restart times $1= t_1 \leq t_2 \leq \dots \leq t_{m+1} = n+1$ such that 
\begin{equation}
  \label{eq:fixed_batches}
  TV\big(\theta_{t_{i}:(t_{i+1}-1)}\big) \leq \frac{TV(\theta_{1:n})}{m} \,,
\end{equation}
for all $1\leq i\leq m$ and  then forms the prediction $\tilde y_t$ for $t \in \{t_i+1,\dots,t_{i+1}\}$
\begin{equation}
  \label{eq:constant_pred}
     \tilde y_t  \eqdef \bar y_{t_i:(t-1)} ~ \text{where} ~ \bar y_{t_i:(t-1)} \eqdef \frac{1}{t - t_i} \sum_{k=t_i}^{t-1} y_k \,.
\end{equation}
We would assume the existence of similar hypothetical forecaster for non-stationary online linear regression (section~\ref{sec:linear_reg}) and non-stationary online kernel regression (section~\ref{sec:kernel_reg}) with slight variation in the prediction function.  Of course this forecaster is not practical since the restart times $t_i$ are unknown. 

In Theorem~\ref{thm:approx_error} stated in Appendix~\ref{app:proof_approx}, we show that for $m \approx n^{1/3}C_n^{2/3}$, this hypothetical forecaster achieves the same optimal error of Proposition~\ref{prp:lower_bound_oned},
\begin{equation}
  \label{eq:reg_hypothetic}
  R_n(\tilde y_{1:n}, \theta_{1:n}) \leq O(n^{1/3}C_n^{2/3}) \,,
\end{equation}
as was already obtained by~\cite{baby2019online}. Of course, it remains to estimate the restart times $t_i$. 
\paragraph{A meta-aggregation algorithm to learn the restart times} 
Contrary to~\cite{baby2019online}, which uses a change point detection method, we propose to do so by using meta-aggregation algorithms from non-stationary online learning such Follow the Leading History (FLH) \cite{hazan2007adaptive} based on exponential weights and presented in Algorithm~\ref{alg:flh}.

\begin{algorithm}[th]
\begin{small}
\KwInput{black box algorithm $\cA$, learning parameter $\eta > 0$}

{\bfseries Init:} $S_0 = \emptyset$

\For{ $t = 1,\dots,n$}{
    Start a new instance of algorithm $\cA$ denoted $\cA_t$ and assign weight $\hat p_t(t) = \frac{1}{t}$.

    Normalize the weight of each expert $j \in [t-1] := \{1,\dots,t-1\}$
    \[
        \textstyle{\hat p_t(j) := \big(1- \frac{1}{t}\big) \frac{p_t(j)}{\sum_{j \in [t-1]} p_t(j)}}
    \]

    Observe $x_t$ and get the prediction $\tilde y_t(i)$ from each black box algorithm $\cA_i, i \in [t-1]$.

    Predict $\hat y_t = \sum_{i \in [t-1]} \hat p_t(i) \tilde y_t(i)$ and observe $y_t \in \R$.

    Update the weights for each $i \in [t-1]$
    \[
        p_{t+1}(i) = p_t(i) \exp\big(-\eta (y_t - \tilde y_t(i))^2\big) \,.
    \]
}
\end{small}
\caption{Follow the Leading History (FLH) \cite{hazan2007adaptive}}
\label{alg:flh}
\end{algorithm}
Basically, FLH is a meta-aggregation procedure that considers a subroutine algorithm, called $\cA$, producing a prediction based on past observations. $\cA$ can be any online learning algorithm that aims at minimizing the static regret, that is the excess cumulative error compared to a fixed parameter. The role of the meta-algorithm is to learn the restarts. To do so, at each round $t\geq 1$, FLH builds a new expert (step 3 of Alg.~\ref{alg:iflh_d_1}) that applies $\cA$ on the sequence of observations $y_t,\dots,y_n$ (that is by not considering the past data before round $t$). This new expert is assigned a weight $1/t$ and the weights of previous experts are normalized so that they sum to 1 (step 4). 
All the experts are then combined using a standard exponentially weighted average algorithm (step 6 of Alg.~\ref{alg:iflh_d_1}). The prediction of FLH is finally obtained (step 8) by forming a convex combination of the expert predictions. The number of active experts grow linearly with time. In Alg.~\ref{alg:iflh_d_1}, we also present IFLH, introduced by \cite{zhang2017dynamic}, which improves the computational complexity by removing experts over time.

In Theorem~\ref{thm:one_d_adaptive}, we show that a cumulative error of optimal order $\tilde O(C_n^{2/3} n^{1/3})$ can be achieved by applying FLH with moving averaged \eqref{eq:constant_pred} as subroutines. 
\begin{theorem} 
\label{thm:one_d_adaptive}
Let $\theta_{1:n} \in \R^{n}$ such that $TV(\theta_{1:n}) \leq C_n$. Assume that $|\theta_t| \leq B$ for all $t\geq 1$. If moving average predictions~\eqref{eq:constant_pred} are used as subroutine of Algorithm~\ref{alg:iflh_d_1}, the cumulative error is upper-bounded as
\begin{small}
\[
   R_n(\hat y_{1:n},\theta_{1:n}) \leq O(n^{1/3} C_n^{2/3}\log^2 n)  \,,
\]
\end{small}
with high probability. 
\end{theorem}
\begin{proof}
First, with probability $1-\delta$, all $|y_t| = |\theta_t + Z_t|$ for $1\leq t\leq n$ are bounded by $C \sqrt{\log n}$ for some constant $C$ depending on $B$ and $\sigma^2$. Thus, $y \mapsto (y - y_t)^2$ are $\alpha$-exp-concave with $\alpha = C'/(\log(n/\delta))$ for some $C' >0$.  Let $m \approx n^{1/3} C_n^{2/3}$ and $t_1,\dots,t_m$ be as defined in \eqref{eq:fixed_batches} and~\eqref{eq:reg_hypothetic} (see also Thm. \ref{thm:approx_error}). From Claim~3.1 of \cite{hazan2007adaptive}, we have for any $i = 1,\dots, m$
\begin{small}
\[
  \sum_{t=t_i}^{t_{i+1}-1} (\hat y_t - y_t)^2  - (\tilde y_t(t_i) - y_t)^2 \leq \frac{3 \log n}{\alpha} \leq O(\log^2 n) \,.
\]
\end{small} \noindent 
Therefore, summing over $i=1,\dots,m$ and using that the subroutines are moving averages (i.e., $\tilde y_t(t_i) = \bar y_{t_i:(t-1)}$) and the definition of $\tilde y_t$ in~\eqref{eq:constant_pred}, we get
\begin{small}
\begin{equation}
  \label{eq:reg_hazan}
    \sum_{t=1}^n (\hat y_t - y_t)^2  - (\tilde y_t - y_t)^2 \leq O(m \log^2 n) \,.
\end{equation}
\end{small}
Thus, because $Z_t = y_t - \theta_t$ is independent of $\hat y_t$
\begin{small}
\begin{align*}
 & R_n (\hat y_{1:n}, \theta_{1:n})  := \sum_{t=1}^n \E\big[(\hat y_t - y_t + y_t - \theta_t)^2\big] \\
  & = \sum_{t=1}^n \E\big[(\hat y_t - y_t)^2 - (y_t - \theta_t)^2\big] \\
  & = \sum_{t=1}^n \E\big[(\hat y_t - y_t)^2 - (\tilde y_t -y_t)^2 \\ 
& \hspace*{2cm} +  (\tilde y_t - y_t)^2 - (y_t - \theta_t)^2\big] \\
& \stackrel{\eqref{eq:reg_hazan}}{\leq} O(m (\log n)^2) + \sum_{t=1}^n \E\big[ (\tilde y_t - y_t)^2 - (y_t - \theta_t)^2\big] \,.
\end{align*}
\end{small}
It only remains to show that the last term corresponds to $R_n(\tilde y_{1:n}, \theta_{1:n}) := \sum_{t=1}^n \E\big[ (\tilde y_t - \theta_t)^2\big]$ and apply Inequality~\eqref{eq:reg_hypothetic}. Expending the squares, it indeed  yields
\begin{small}
\begin{align*}
    \E\big[& (\tilde y_t - y_t)^2 - (y_t - \theta_t)^2\big]  = \E\big[\tilde y_t^2  + 2(\theta_t - \tilde y_t) y_t - \theta_t^2 \big]\\
    & =  \E\big[\tilde y_t^2  + 2(\theta_t - \tilde y_t) (\theta_t + Z_t) - \theta_t^2 \big]\\
    & = \E\big[ (\tilde y_t - \theta_t)^2\big] \,,
\end{align*} 
\end{small}
where the last equality is because $\E[Z_t] = 0$ and $Z_t$ is independent from $\tilde y_t$ and $\theta_t$. 
\end{proof}

%
%

\vspace{-1mm}
\section{Non-Stationary Online Regression}\label{sec:regression}
\vspace{-2mm}

In this section, we discuss more general problem of non-stationary online regression. 
We consider the following problem :
\begin{equation}
y_t = g_t(x_t) + Z_t 
\label{eq:generalized_curve_fitting}
\end{equation}
where $g_t : \mathbb{R}^d \rightarrow \mathbb{R}$ is a non-linear function and $Z_t$ be independent $\sigma$-subGaussian random variables in one dimension with $\E[Z_t] = 0$. Similar to the previous section, the goal in this section would be to track the sequence of $g_t$ with $\hat{g}_t$ for all $t$ such that $\hat{y}_t = \hat{g}_t(x_t)$ to minimize the expected cumulative error $  R_n(\hat y_{1:n},\theta_{1:n})$ with respect to the unobserved output $g_t$ after $n$ time steps which we define as follow: 
\begin{align}
   R_n(\hat y_{1:n},g_{1:n}) =\sum_{t=1}^n \bbE \left[(\hat{y}_t - {g}_t(x_t))^2\right] \notag \\
   =  \sum_{t=1}^n \bbE \left[(\hat{g}_t(x_t) - {g}_t(x_t))^2\right].  \label{eq:gen_non-station}
\end{align} 
However, we need to remember that we observe $g_t(x_t)$ only after perturbed through some noise variable $Z_t$. Hence, we need to decompose our regret in terms of the observed response $y_t$. Bias-variance decomposition directly provides the decomposition in terms of the observed variable $y_t$. Proof is given in Appendix~\ref{ap:linear}.
\begin{lemma} \label{lem:bias_variance_square}
For any sequence of functions $\tilde{g}_t : \mathbb{R}^d\rightarrow \mathbb{R}$ for $t \in [n]$ independent of $Z_t$ for all $t$, the cumulative error~\eqref{eq:gen_non-station} can be decomposed as follows:
\begin{small}
\begin{align*}
R_n(\hat y_{1:n},g_{1:n})= \sum_{t=1}^n \bbE \left[(\hat{y}_t - {y}_t)^2 - (\tilde{g}_t(x_t) - {y}_t)^2 \right] \notag \\
 + \sum_{t=1}^n \bbE\left[ (\tilde{g}_t(x_t) - {g}_t(x_t))^2\right].
\end{align*}
\end{small}
\end{lemma}

\subsection{Non-stationary Linear Regression} \label{sec:linear_reg}
\vspace{-1mm}
\begin{algorithm}[th]
\begin{small}
\KwInput{black box algorithm $\cA$, learning parameter $\eta > 0$}

{\bfseries Init:} $S_0 = \emptyset$

\For{ $t = 1,\dots,n$}{
    Start a new instance of algorithm $\cA$ denoted $\cA_t$ and assign weight $\hat p_t(t) = \frac{1}{t}$.

    Define its ending time as 
        $\tau_t := t + 2^k$ where $k := \min \{k\geq 0 \ \text{s.t}\  c_k >0 \}$ and $t := \sum_{i=1}^\infty c_k 2^k$ is the binary representation of $t$. 

    Define the set of active experts $S_t := \{1\leq i \leq t: \ \tau_i > t \}$

    Normalize the weight of each active expert $j \in S_t \setminus \{t\}$
    \[
        \hat p_t(j) := \big(1- \frac{1}{t}\big) \frac{p_t(j)}{\sum_{j \in S_t \setminus \{t\}} p_t(j)}
    \]

    Observe $x_t$ and get the prediction $\tilde y_t(i)$ from each black box algorithm $\cA_i, i \in S_t$.

    Predict $\hat y_t = \sum_{i \in S_t} \hat p_t(i) \tilde y_t(i)$ and observe $y_t \in \R$.

    Update the weights for each $i \in S_t$
    \[
        p_{t+1}(i) = p_t(i) \exp\big(-\eta (y_t - \tilde y_t(i))^2\big) \,.
    \]
}
\end{small}
\caption{IFLH  -- Improved Following the Leading History (binary base) \cite{zhang2017dynamic}}
\label{alg:iflh_d_1}
\vspace{-2mm}
\end{algorithm}
In Lemma~\ref{lem:bias_variance_square}, we provided the general bias-variance decomposition result for squared loss while computing expected cumulative error. In this section, we will specifically discuss the result for linear predictor $\theta_t$ for all $t$ \textit{i.e.} we assume that $g_t$ is linear function for all $t$. Hence, the problem can be formulated as follows. At each step $t\geq 1$, the learner observes $x_t \in \R^d$, predicts 
$\hat y_t = x_t^\top \hat \theta_t$ and observes
\begin{equation}
      y_t = x_t^\top \theta_t + Z_t \,
      \label{eq:linear_noise}
\end{equation}
where  $Z_t$ be independent $\sigma$-subGaussian zero mean random variable.  We assume $\theta_1,\dots, \theta_n \in \R^d$ such that $TV(\theta_{1:n}) =  \sum_{t=2}^n \|\theta_t - \theta_{t-1}\|_1 \leq  C_n$ and $\|\theta_t\| \leq B$ for all $t\geq 1$. 
The goal is to control the cumulative error with respect to the unobserved outputs $\tilde{y}_t = x_t^\top \theta_t = y_t - Z_t.$ We substitute $g_t(x_t) $ with $ x_t^\top \theta_t$ in Equation~\eqref{eq:gen_non-station} and denote the prediction function $\hat{g}_t(x_t) = x_t^\top \hat{\theta}_t = \hat{y}_t $. Hence, the expected cumulative error $R_n(\hat y_{1:n},g_{1:n})$ can be written as 
\[
     \sum_{t=1}^n \E\Big[(\hat y_t - \tilde y_t)^2 \Big] = \sum_{t=1}^n \E\Big[\big((\hat \theta_t - \theta_t)^\top x_t\big)^2 \Big] \,.
\]
\vspace{-2mm}
\paragraph{Hypothetical forecaster} 
We consider an hypothetical forecaster which similar to that of  1-dimensional case. It computes a sequence of restart times $1= t_1 \leq t_2 \leq \dots \leq t_{m+1} = n+1$ 
for all $1\leq i\leq m$ as in equation~\eqref{eq:fixed_batches} and then forms the prediction $\tilde y_t$ for $t \in \{t_i+1,\dots,t_{i+1}\}$
\begin{equation}
  \label{eq:constant_pred_d}
     \tilde y_t  \eqdef x_t^\top \bar{\theta}_{t} \,,
\end{equation}
where $\bar \theta_t = \bar \theta_{t_j:(t_{j+1}-1)}$ for $t_j \leq t < t_{j+1}$ and $\bar \theta_{t_j:(t_{j+1}-1)}  = \frac{1}{t_{j+1} -t_j} \sum_{t=t_j}^{t_{j+1}-1} \theta_t$. Below in Lemma~\ref{lem:approx_least_square}, we show that the cumulative  error can be controlled with respect to this hypothetical forecaster. 
\begin{lemma}[Adaptive Restart in  $d$-dimension]
\label{lem:approx_least_square}
Let $X,B>0$. Assume that $\|x_t\|\leq X$ and $\|\theta_t\| \leq B$ for all $t\in [n]$. Then, there exists a sequence of restarts $1= t_1<\dots<t_m = n+1$ such that
\begin{small}
\begin{align*}
   \sum_{t=1}^n (x_t^\top \bar \theta_t - x_t^\top  \theta_t )^2  &=  \sum_{j=1}^m \sum_{t=t_j}^{t_{j+1}-1} \big((\bar \theta_{t_j:(t_{j+1}-1)}  - \theta_t )^\top x_t\big)^2 \\
   &\leq X^2 n \Big(\frac{C_n}{m}\Big)^2 + 4 X^2 B^2 m \,,
\end{align*}
\end{small}
where $\bar \theta_t := \bar \theta_{t_j:(t_{j+1}-1})$ for   $t_j \leq t \leq t_{j+1} - 1$  {and} $ \bar \theta_{t_j:(t_{j+1}-1)}  = \frac{1}{t_{j+1} -t_j} \sum_{t=t_j}^{t_{j+1}-1} \theta_t$.
\end{lemma}


However, this forecaster cannot be computed and is only useful for the analysis since both the restart times $t_i$ and the parameters $\theta_t$ are unknown.  We use meta algorithm \emph{Improved Following the Leading History} (IFLH, Algorithm~\ref{alg:iflh_d_1}) \cite{zhang2017dynamic} to efficiently learn the restart time which is computationally more efficient than FLH presented in Algorithm~\ref{alg:flh}.  To reduce the computation complexity, there is also an associated ending time for each expert in IFLH which tells that that particular  expert will no longer active after its ending time. As we only have the access to the noisy gradient, we will utilize the result presented in \cite[Theorem 1]{zhang2017dynamic} with a probabilistic upper bound on the gradient to get the final upper bound on expected cumulative loss. We provide below an upper bound on the expected cumulative error.


\begin{theorem} \label{thm:linear_non-station_reg}
Let $n, m \geq 1$, $\sigma >0$, $B>0$, $X>0$, and $C_n >0$. Let $\theta_1,\dots,\theta_n$ such that $TV(\theta_{1:n}) \leq C_n$ and $\|\theta_t\|\leq B$. Assume that $\|x_t\|\leq X$ for all $t\geq 1$. Then, Alg.~\ref{alg:iflh_d_1} \cite{zhang2017dynamic} with Online Newton Step \cite{hazan2007logarithmic} as subroutine and well-tuned learning rate $\eta>0$  satisfy
\[
  R_n(\hat y_{1:n}, \theta_{1:n}) \lesssim  d^{1/3}n^{1/3} C_n^{2/3} (X^2\sigma^2 B + X^2 B^2)^{1/3}  \,,
\]
with high probability.
\end{theorem}
\paragraph{Discussion:}The result presented in Theorem~\ref{thm:linear_non-station_reg} provides an upper bound on the expected cumulative error of Alg.~\ref{alg:iflh_d_1} for non-stationary online linear regression. This generalizes the result of \citet{baby2019online} which only works for one dimensional problem. Our algorithm is adaptive to the noise parameter $\sigma$ which means we do not need to know the variance $\sigma$, which is not correct for the algorithm presented in \citet{baby2019online}. While implementing the algorithm, all we need to know is the maximum value of $y_t$ observed so far. 
\paragraph{On Lower Bound:} The lower bound presented in \citet{baby2019online} can be extended easily for general $d$-dimension by considering the problem of $d$-independent variables. This will simply add an extra multiplicative factor of $d$ in the lower bound (Proposition~\ref{prp:lower_bound_oned}). Our upper-bound is thus optimal in $n, d$, and $C_n$. However, the dependence in $\sigma$ is worse than the one of \citet{baby2019online}. This may be due to fact that our algorithm also adapt to the noise parameter and we do not need to know $\sigma$ in our algorithm. It is an interesting question to know whether our dependence in $\sigma$ is optimal in our case and we leave it for future work.



\subsection{Non-stationary Kernel Regression} \label{sec:kernel_reg}
In this section, we  consider the case of  non-stationary online kernel regression.  For  the input space $\mathcal{X}$  and a positive
definite kernel function $\mathcal{K}:\mathcal{X}\times \mathcal{X} \rightarrow \mathbb{R}$,  we denote the RKHS associated with $\mathcal{K}$ as $\mathcal{H}$.  We further denote the associated feature map $\phi:\mathcal{X}\rightarrow \mathcal{H}$, such that $\mathcal{K}(x,x') = \langle \phi(x), \phi(x') \rangle_{\mathcal{H}}$. With slight abuse of notation, we write that $\mathcal{K}(x,x')   = \phi(x) ^\top \phi(x') .$ In this section, we assume that the functions $g_t$ lie in some RKHS $\mathcal{H}$ corresponding to the kernel $\mathcal{K}$  for all $t$. At each step $t\geq 1$, the learner observes $x_t \in \R^d$, predicts 
$\hat y_t = \phi(x_t)^\top \hat\theta_t$ and observes
\begin{align}
      y_t = \phi(x_t)^\top \theta_t + Z_t \,,  \label{eq:RKHS_well_specified}
\end{align}
where  $Z_t$ be independent $\sigma$-subGaussian zero mean random variable.  The case we consider comes under well specified case as the optimal functions $\theta_1,\dots, \theta_n \in \mathcal{H}$ lie in the same RKHS $\mathcal{H}$ corresponding to the kernel $\mathcal{K}$ where we consider our hypothesis space. We define $K_{nn}$ as $(K_{nn})_{i,j} = \langle\phi(x_i), \phi(x_j)\rangle$ and $\lambda_k(K_{nn})$ denotes the $k$-th largest eigenvalue of $K_{nn}$. Time dependent effective dimension $d_{eff}(\lambda,s,r) $ is defined as follows, 
\begin{align*}
d_{eff}(\lambda,s,r) = Tr(K_{s-r,s-r}(K_{s-r,s-r}+\lambda I)^{-1})\,.
\end{align*}
We also assume  that $TV(\theta_{1:n}) = \sum_{t=2}^n \|\theta_t - \theta_{t-1}\|_{\mathcal{H}} \leq C_n$. The goal is to control the cumulative error with respect to the unobserved outputs $\tilde{y}_t = \phi(x_t)^\top \theta_t = y_t - Z_t.$ We substitute $g_t(x_t) $ with $ \phi(x_t)^\top \theta_t$ in Equation~\eqref{eq:gen_non-station} and denote the prediction function with $\hat{\theta}_1,\cdots , \hat{\theta}_n$ such that $\hat{g}_t(x_t) = \phi(x_t)^\top \hat{\theta}_t = \hat{y}_t $. Hence, the expected cumulative error $R_n(\hat y_{1:n}, \theta_{1:n}) $ can be written as 
\begin{align*}
     R_n(\hat y_{1:n}, \theta_{1:n})   &= \sum_{t=1}^n \E\Big[\big((\hat \theta_t - \theta_t)^\top \phi(x_t) \big)^2 \Big] \,.
\end{align*}

 For our analysis, we consider a similar hypothetical forecaster as in linear regression  (see Equation~\eqref{eq:fixed_batches}). The prediction $\tilde y_t$ for $t \in \{t_i+1,\dots,t_{i+1}\}$ is simply given as $\tilde y_t  \eqdef \phi(x_t)^\top \bar{\theta}_{t}$ where $\bar \theta_t = \bar \theta_{t_j:(t_{j+1}-1)}$ for $t_j \leq t < t_{j+1}$. In the result  given below in Lemma~\ref{lem:approx_least_square_rkhs}, we show that the expected cumulative error can be controlled with respect to this hypothetical forecaster given the adaptive restart. 
\begin{lemma}[Adaptive Restart in RKHS]
\label{lem:approx_least_square_rkhs}
Let $B, \kappa >0$. Assume that $\|\phi(x_t) \|^2 \leq \kappa^2$, and $\|\theta_t\|_{\mathcal{H}} \leq B$ for all $t$. Then, there exists a sequence of restarts $1= t_1<\dots<t_m = n+1$ such that
\begin{small}
\begin{align*}
   &\sum_{t=1}^n (\phi(x_t)^\top \bar \theta_t - \phi(x_t)^\top  \theta_t )^2  \\
   &\qquad \qquad =  \sum_{j=1}^m \sum_{t=t_j}^{t_{j+1}-1} \big((\bar \theta_{t_j:(t_{j+1}-1)}  - \theta_t )^\top \phi(x_t)\big)^2  \\
   &\qquad \qquad \leq \kappa^2 n \Big(\frac{C_n}{m}\Big)^2 + 4\kappa^2 B^2 m\,,
\end{align*}
\end{small}
where $\bar \theta_t := \bar \theta_{t_j:(t_{j+1}-1)}$ for $t_j \leq t < t_{j+1}$, $C_n \geq \sum_{t=2}^n \|\theta_t - \theta_{t-1} \|_{\mathcal{H}}$, and
\[
    \bar \theta_{t_j:(t_{j+1}-1)}  = \frac{1}{t_{j+1} -t_j} \sum_{t=t_j}^{t_{j+1}-1} \theta_t.
\]
\end{lemma}
As we have discussed previously, it is not possible to compute this forecaster and it  will be only useful in the analysis of the algorithm. One simply has to use a meta algorithm as in \cite{zhang2017dynamic}  to learn these restart times. However, one cannot use online newton step as the black box subroutine in this meta algorithm like it was done for linear regression as the convergence of online newton step is not known for tracking prediction functions in RKHS. Hence, we use Kernel-AWV as the black box online learner \cite{gammerman2012line} (see also \cite{jezequel2019efficient}) as subroutine in Alg.~\ref{alg:iflh_d_1} to estimate the prediction function. Kernel-AWV depends on a regularization parameter $\lambda >0$. Note that other subroutines designed for Online Kernel Regression such as Pros-N-Kons \cite{calandriello2017second} or PKAWV \cite{jezequel2019efficient} can be used. 
Below, we have the following theorem regarding the adaptive regret of least square in when the predictor function lies in RKHS.
\vspace{-1mm}
 \begin{theorem} \label{thm:zhang_rkhs}
Let $\lambda >0$. For online kernel regression with square loss if for all $i \in [n]$, $y_i \in [-Y,Y]$ then for Algorithm~\ref{alg:iflh_d_1} with Kernel-AWV  \cite{gammerman2012line} with regularization parameter $\lambda$ as subroutine,  we have
\begin{small}
\begin{align*}
    \sum_{t=r}^s f_t(\theta_t) - \sum_{t=r}^s f_t(u) \leq 8Y^2({p+2})\log n +  \lambda p \| \theta\|^2 \\
    + Y^2p d_{eff}(\lambda,s-r) \log \left( e + \frac{en\kappa^2}{\lambda}\right).
\end{align*}
\end{small}
where $ [r,s] \subseteq [n] $, $p \leq \lceil \log_2(s-r+1) \rceil + 1$ and $f_t(\theta) = (y_t - \phi(x_t)^\top \theta)^2$. 
\end{theorem}
\vspace{-1mm}
With Theorem~\ref{thm:approx_error} and Lemma~\ref{lem:approx_least_square_rkhs}, we have the expression for upper bound on both the independent error terms which after combining together bound the overall expected cumulative error. Below, we provide our final bound on the expected cumulative error assuming the capacity condition, i.e., that the effective dimension satisfies $d_{eff}(\lambda, n) \leq ({n}/{\lambda})^\beta$ for $\beta \in (0,1)$. The proof is given in Appendix~\ref{ap:kernel}.
\vspace{-1mm}
\begin{theorem} \label{thm:kernel_non_station_reg}
Let $n, m \geq 1$, $\sigma >0$, $B>0$, $\kappa >0$, and $C_n >0$. Let $\theta_1,\dots,\theta_n$ such that $TV(\theta_{1:n}) \leq C_n$ and $\|\theta_t\|_{\mathcal{H}} \leq B$ for all $t\geq 1$. Assume also that $\|\phi(x_t)\|_{\mathcal{H}} \leq \kappa$ for $t\geq 1$.  Then , for well chosen $\eta>0$, Alg.~\ref{alg:iflh_d_1} with Kernel-AWV using $\lambda = ({n}/{m})^{\frac{\beta}{\beta +1}}$ satisfies 
\begin{small}
\begin{align*}
  R_n(\hat y_{1:n}, \theta_{1:n}) \leq  \tilde{\mathcal{O}} \left( C_n^{\frac{2(\beta+1)}{2\beta+3}} n^{\frac{1}{2\beta+3}} \left( \sigma^2 \log \frac{n}{\delta}   + B^2\kappa^2  \right)\right. \\
  +  \left.C_n^{\frac{2}{2\beta+3}} n^{\frac{2\beta+1}{2\beta+3}} B^{\frac{4(\beta+1)}{2\beta+3}}\kappa^{\frac{2}{2\beta+3}}\right)  \,.
\end{align*}
\end{small}
with probability at least $1-\delta$. 
\end{theorem}
\vspace{-3mm}
\paragraph{Discussion:} To the best of our knowledge, this work is the first extension of non-stationary online regression to non-stationary kernel regression. After carefully looking at the bound on the expected cumulative regret term presented in Theorem~\ref{thm:kernel_non_station_reg} and comparing it with that of non-stationary online linear regression (Theorem~\ref{thm:linear_non-station_reg}),  we find that as $\beta \rightarrow 0$, we have  $\lambda \rightarrow \mathcal{O}(1/d)$  and we would have the similar dependence of $C_n$ and $n$ in the expected cumulative error bound for linear and kernel part.  However, we have a slightly worse dependent on the variance of the noise $\sigma$ in the expected cumulative error bound for non-stationary online kernel regression than that of non-stationary online linear regression part. This artefact arises due to difficulty in simultaneously choosing optimal number of restart time  $m$ and regularization parameter $\lambda$. We believe that the dependence in $\sigma$ in Theorem~\ref{thm:kernel_non_station_reg} can be improved further. 

As discussed in \citep{jezequel2019efficient}, the per round space and time complexities is of order $\mathcal{O}(n^2)$ for each prediction sequence corresponding to different start times. However, the method can be made computationally more efficient by the use of Nyström approximation \citep{jezequel2019efficient}. 

It is also worth pointing out that the optimal learning rate $\eta$ only depends on $B, \kappa, \delta$, and $n$ and can be optimized using standard calibration techniques (e.g., doubling trick). The regularization parameter of $\lambda$ on the other hand depends on the regularity of the Kernel. It can be calibrated by starting at each time steps $t$ in Alg.~\ref{alg:iflh_d_1} several new instances of Kernel-AWV, each run with a different parameter $\lambda$ in a logarithmic grid.

\begin{figure*}[t]
    \begin{subfigure}[h]{0.32\textwidth}
        \centering
        \includegraphics[width=\textwidth]{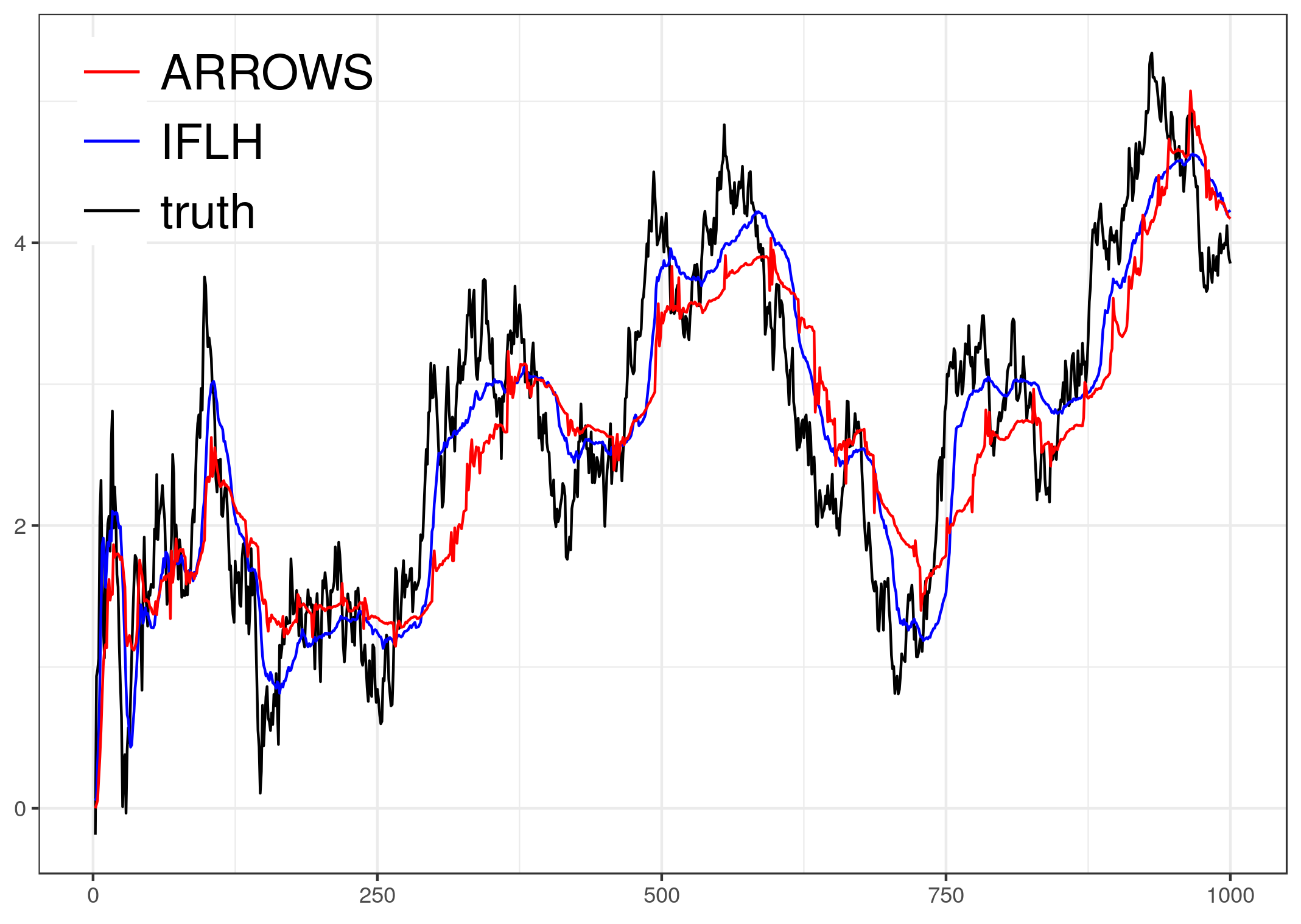}
        \caption{Soft shifts}
        \label{figs:1d-pred-ss} 
    \end{subfigure}~
    \begin{subfigure}[h]{0.32\textwidth}
        \centering
        \includegraphics[width=\textwidth]{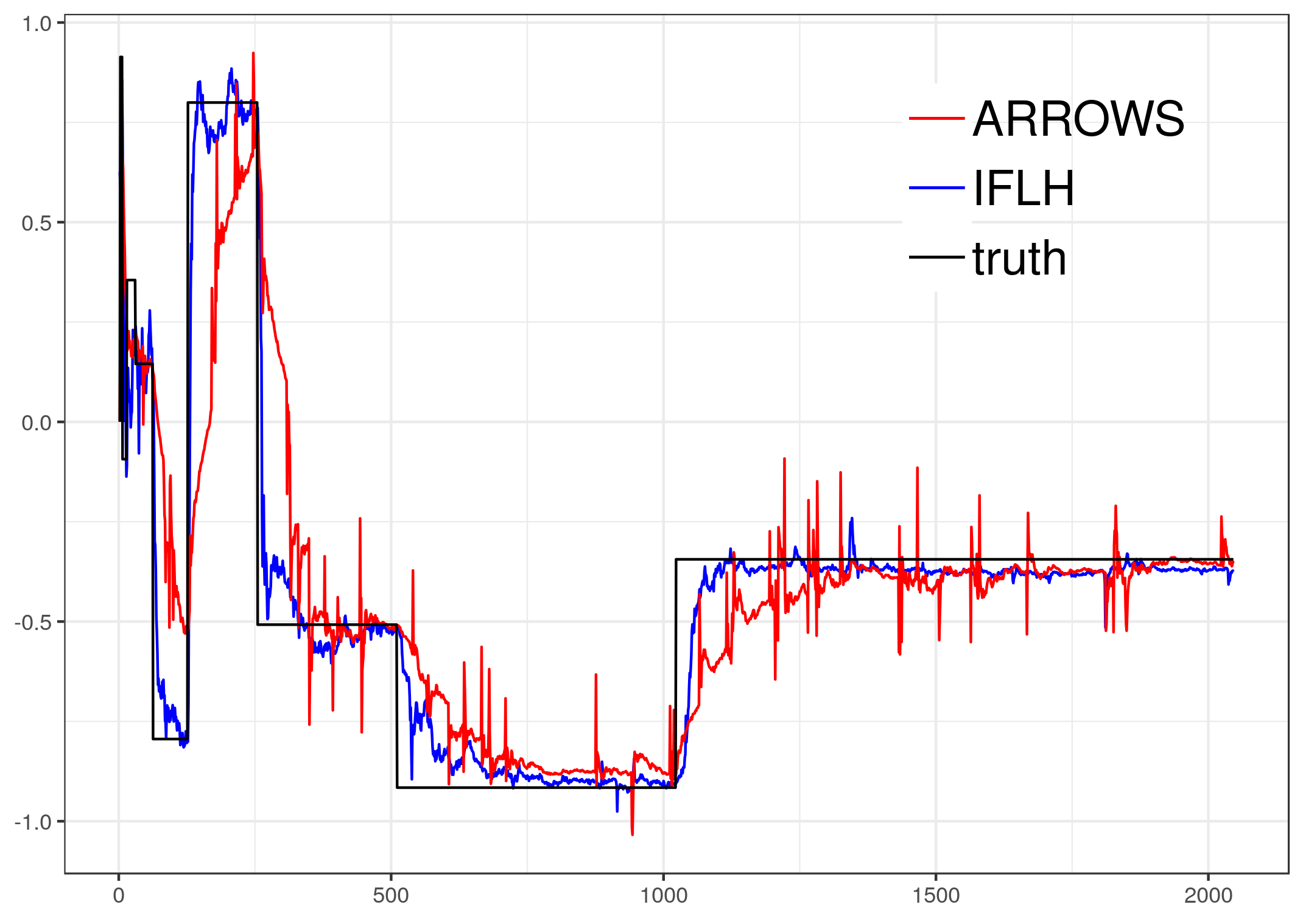}
        \caption{Hard shifts ($m_i = 2^i$, {\small $1 \! \leq \! i\! \leq\!  10$})}
        \label{figs:1d-pred-hs} 
    \end{subfigure}~
    \begin{subfigure}[h]{0.32\textwidth}
        \centering
        \includegraphics[width=\textwidth]{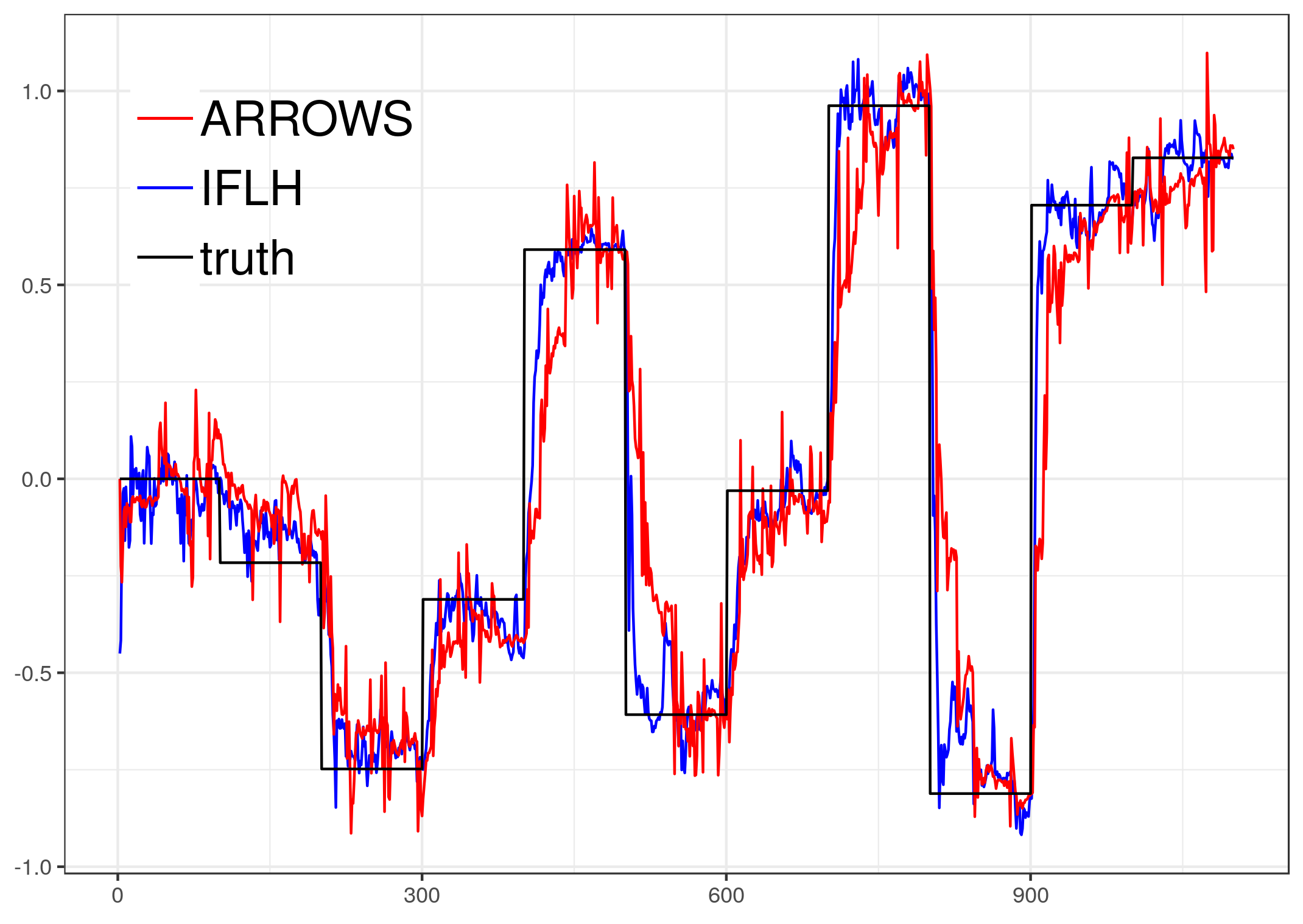}
        \caption{Hard shifts ($m_i = 100i$, {\small $1 \! \leq\! i\!\leq \! 10$})}
        \label{figs:1d-pred-hslog} 
    \end{subfigure}%
    \vspace{-2mm}
    \caption{Examples of predictions obtained by the two considered algorithms (ARROWS in red and IFLH in blue) together with the one dimensional time-series to be predicted.}
    \label{figs:1d-pred} 
\end{figure*}

\begin{figure*}[t]
    \begin{subfigure}[h]{0.32\textwidth}
        \centering
        \includegraphics[width=\textwidth]{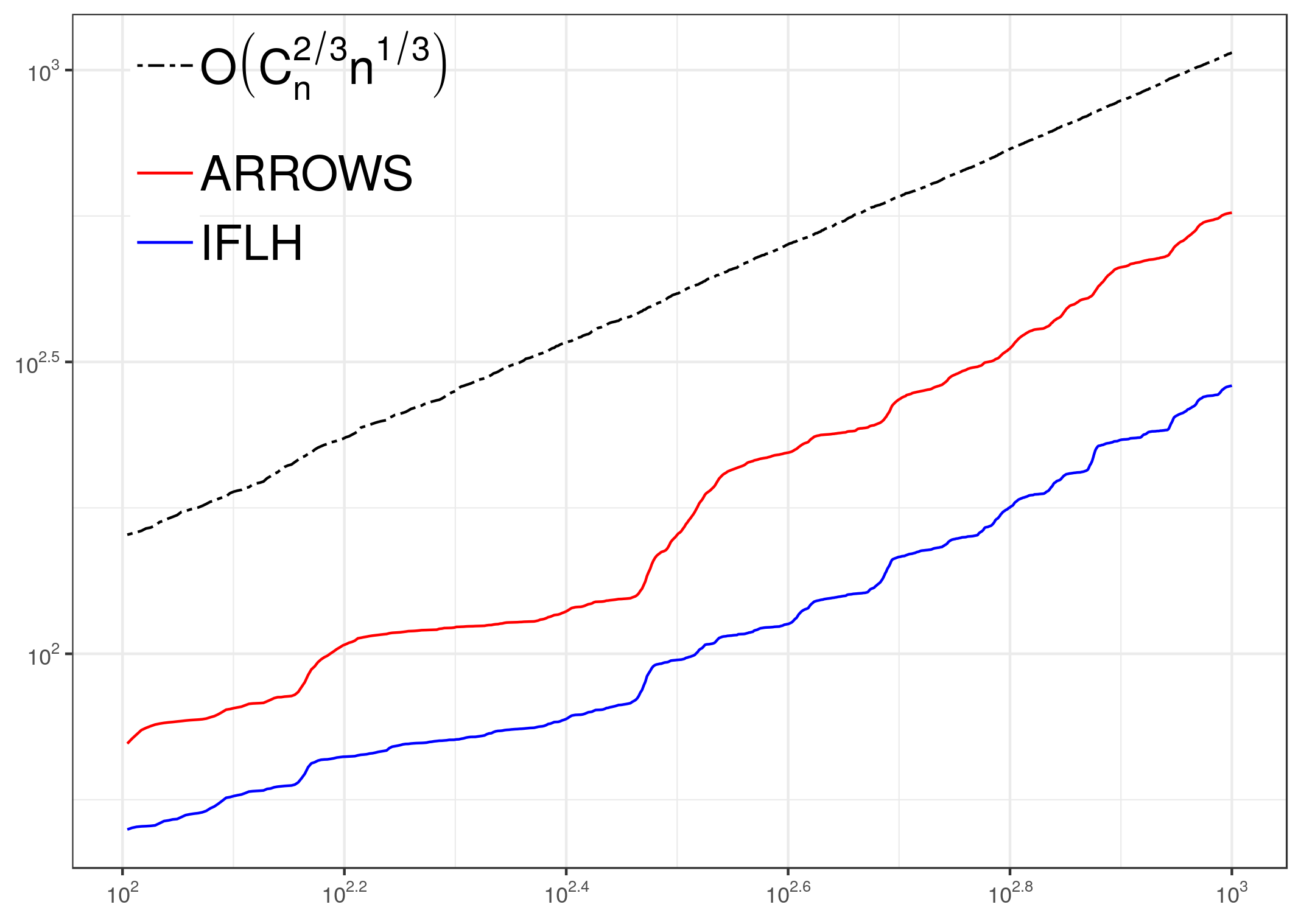}
        \caption{Soft shifts}
        \label{figs:1d-regret-ss} 
    \end{subfigure}~
    \begin{subfigure}[h]{0.32\textwidth}
        \centering
        \includegraphics[width=\textwidth]{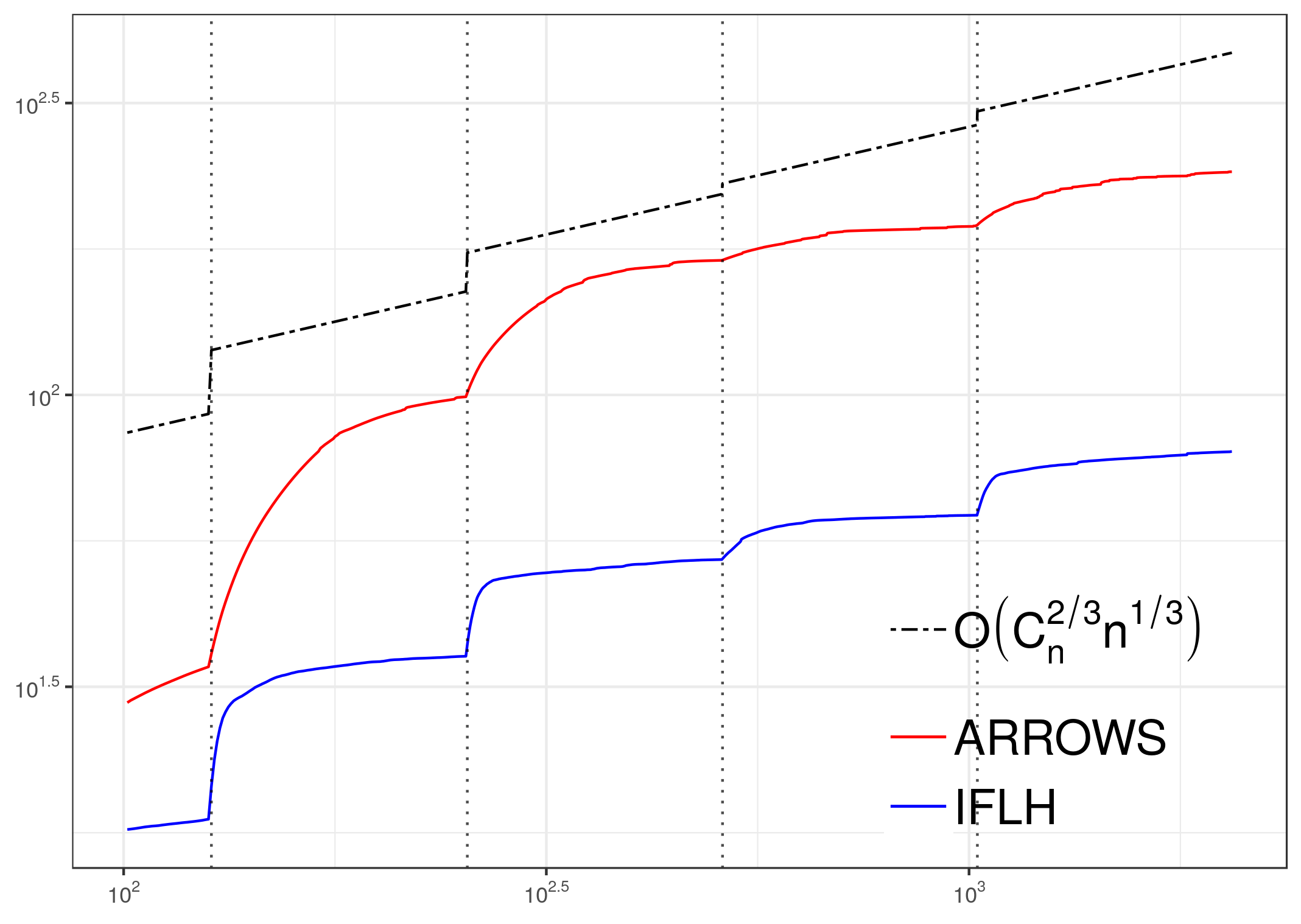}
        \caption{Hard shifts ($m_i = 2^i$, {\small $1 \! \leq \! i\! \leq\!  10$})}

        \label{figs:1d-regret-hs} 
    \end{subfigure}~
    \begin{subfigure}[h]{0.32\textwidth}
        \centering
        \includegraphics[width=\textwidth]{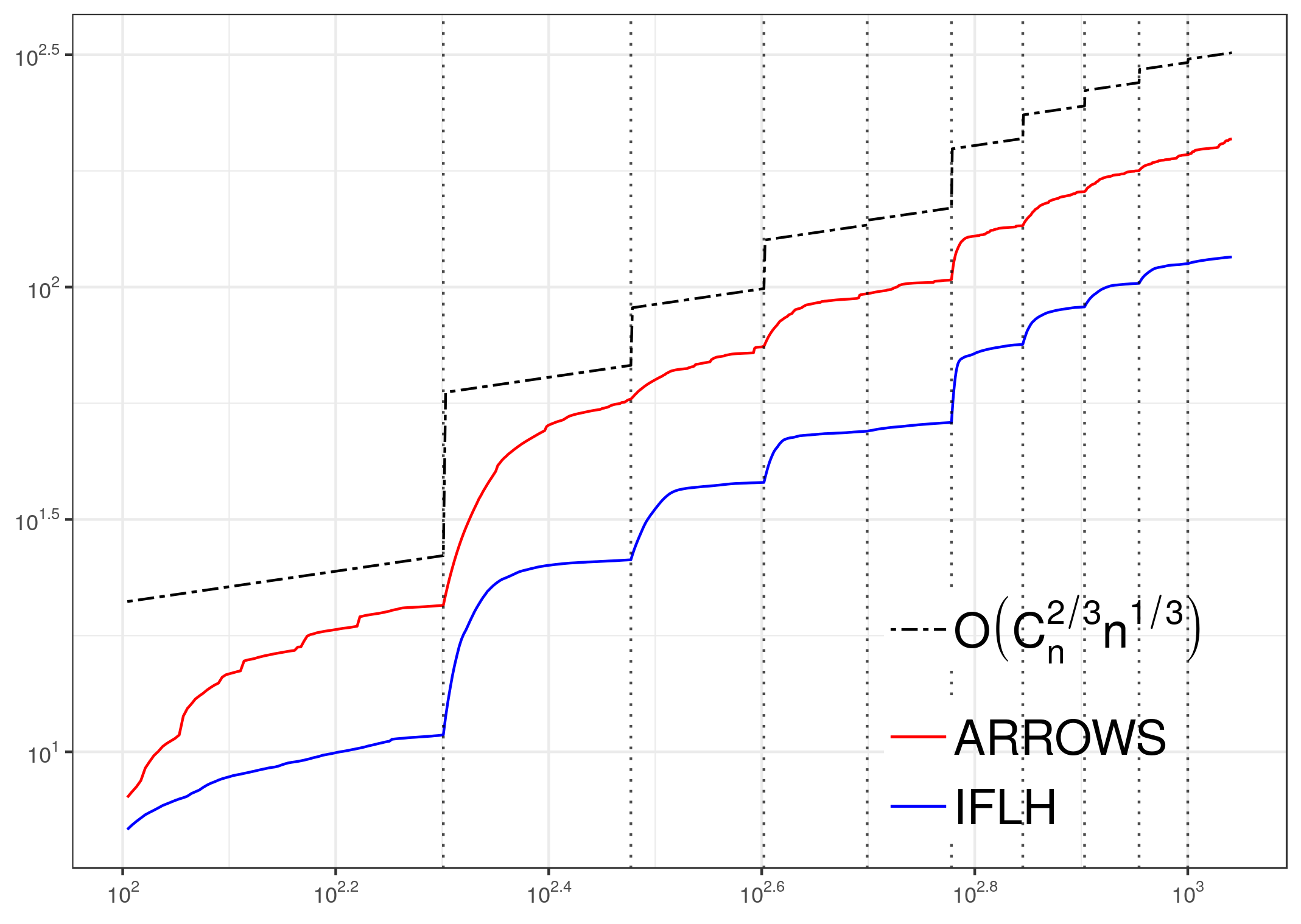}
        \caption{Hard shifts ($m_i = 100i$, {\small $1 \! \leq\! i\!\leq \! 10$})}
        \label{figs:1d-regret-hslog} 
    \end{subfigure}
    \vspace{-2mm}
    \caption{Cumulative errors suffered in average over 10 runs by the two algorithms (ARROWS in red, IFLH in blue) together with the upper bound of order $O(n^{1/3} C_n^{2/3})$ for one dimensional time-series.}
    \vspace{-1mm}
    \label{figs:1d-regret}    
\end{figure*}

\section{Experiments}
\label{sec:exp}
\vspace{-3mm}
In this section, we evaluate our results on empirical simulations. We compare the theoretical bound with the performance of ARROWS \citep{baby2019online} (wherever possible (1 dimension, no input)) and the procedure analyzed here, i.e., IFLH \citep{zhang2017dynamic} with different subroutines (Online Newton Step \cite{hazan2007logarithmic}, OGD \cite{zinkevich2003online}, or Azoury-Warmuth-Vovk forecaster \cite{vovk2001competitive,azoury2001relative}), and online gradient descent with fixed restart \cite{besbes2015non}. We test the algorithms on two different settings. The first one involves a non-stationary time series with continuous small changes in distribution which we call soft shifts. We use decaying innovation variance in order to observe how the algorithms react to a smooth change in the total variation. The second one involves hard and abrupt changes in distribution at well separated time intervals, we call the hard shifts.
\vspace{-3mm}
\subsection{ Data Generation }
\vspace{-3mm}
Before presenting the experimental results and plots, we quickly here discuss the data generation process. Details of data generation process in the setting of soft shifting and hard shifting is given below.
\vspace{-2mm}
\paragraph{Soft Shifts:} We let $\theta_1,\dots,\theta_n$ be a multivariate random walk with exponential decaying variance. We set,  $\theta_t = \theta_{t-1} + \epsilon_t$ with $\epsilon_t \sim  \mathcal{N}(0, t^{-\alpha} I_d)$ multivariate normal. The total variation of this time series is $TV = \sum_{t=2}^n ||\theta_{t} - \theta_{t-1}||_1 = \sum_{t=2}^n ||\epsilon_t||_1$.
\vspace{-2mm}
\paragraph{Hard Shifts:} For generating the data used in hard shifts mechanism, we split the time series $\theta_1, ..., \theta_n \in \R^d$ into $M$ chunks such that $m_i$ is the index of the start of the $i^{\text{th}}$ chunk. At the start of each new chunk, all coordinates $\theta_{m_i}(k)$ for $1\leq k\leq d$ are sampled from independent Rademacher distributions. The values of $\theta_t$ are then constant within a chunk. The total variation of the decision vector $\theta_{1:n}$ is  $TV = \sum_{t=2}^n ||\theta_{t} - \theta_{t-1}||_1 = \sum_{i=2}^{M} ||\theta_{m_i} - \theta_{m_{i-1}}||_1$.
\begin{figure*}[t]
 \centering
    \begin{subfigure}[t]{0.32\textwidth}
        \centering
        \includegraphics[width=\textwidth]{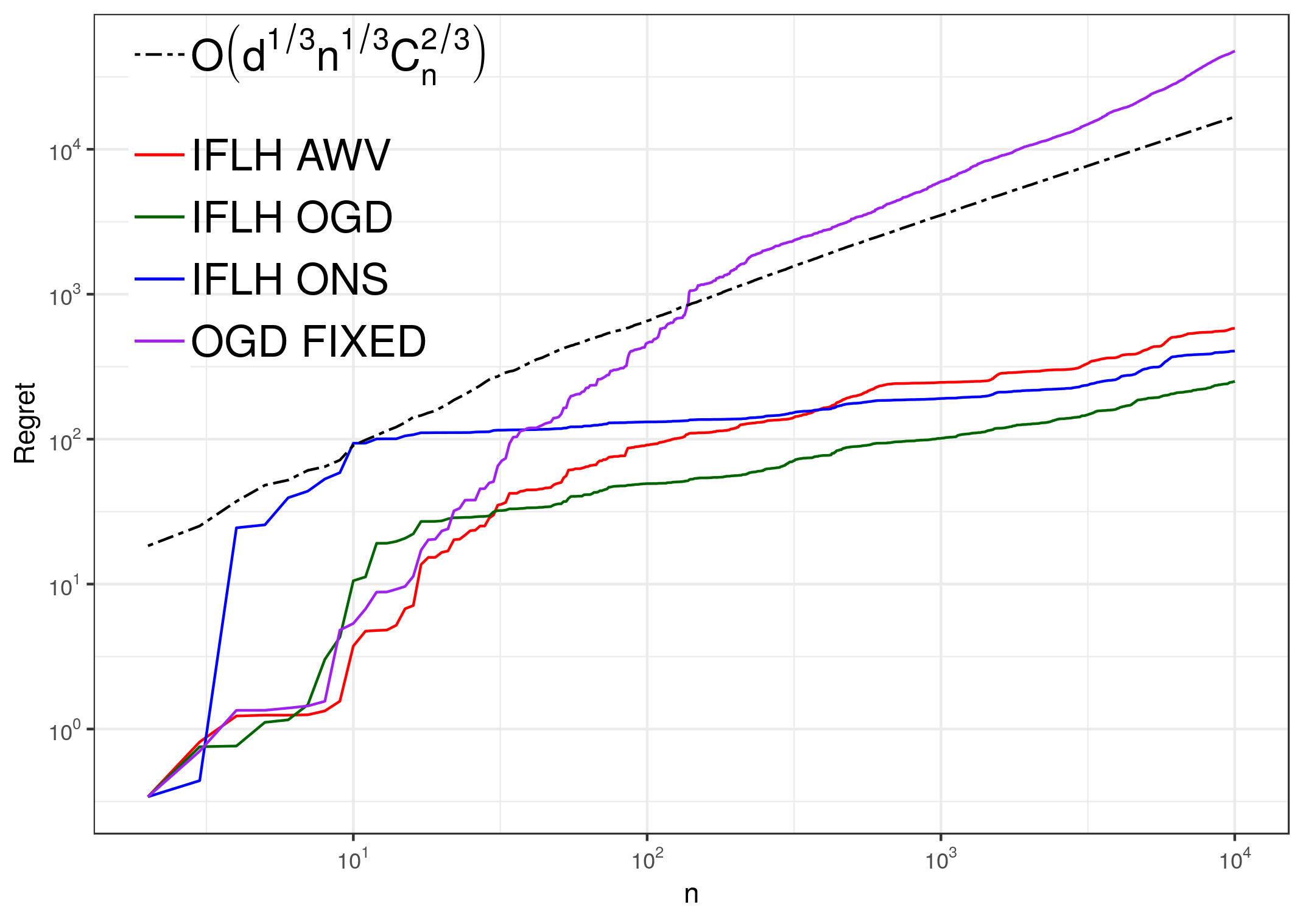}
        \caption{$\alpha = 1$ and $d = 2$}
 \label{fig:lr-ss1d2}   
    \end{subfigure}~
    \begin{subfigure}[t]{0.32\textwidth}
        \centering
        \includegraphics[width=\textwidth]{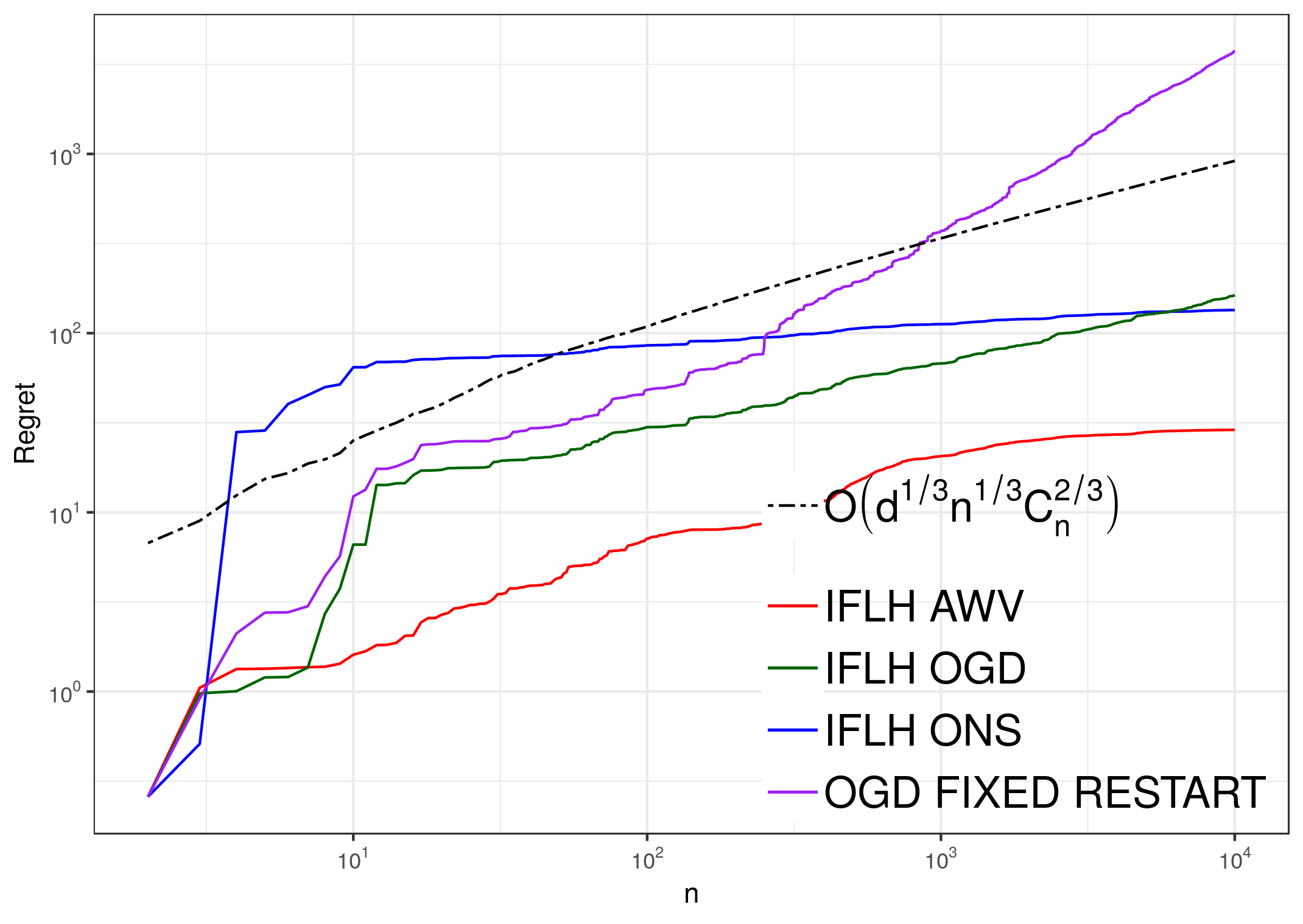}
        \caption{$\alpha = 2$ and $d = 2$}
         \label{fig:lr-ss2d2}
    \end{subfigure}~
    \begin{subfigure}[t]{0.32\textwidth}
        \centering
        \includegraphics[width=\textwidth]{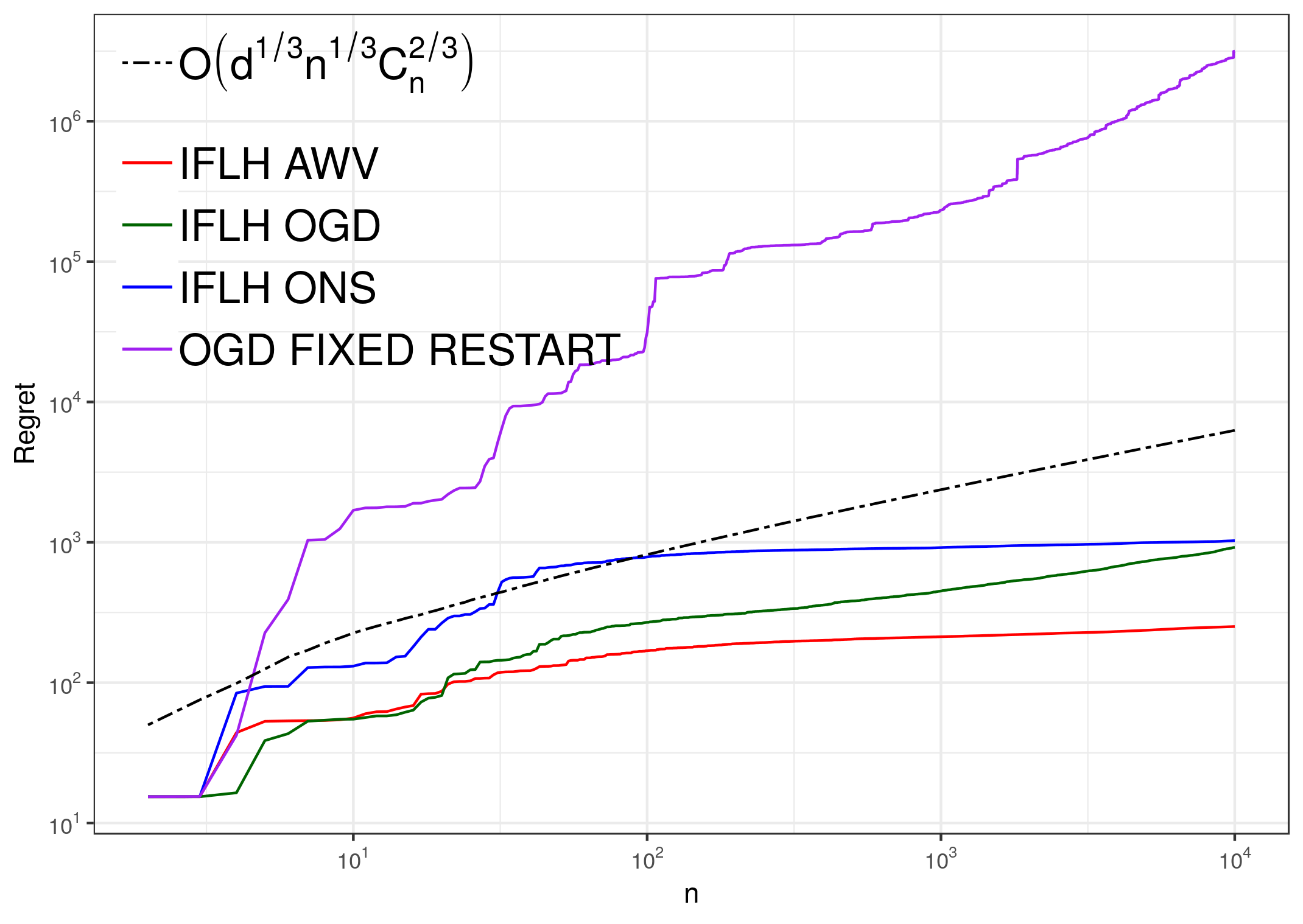}
        \caption{ $\alpha = 2$ and $d = 10$}
        \label{fig:lr-ss2d10}
    \end{subfigure}
    \vspace{-2mm}
    \caption{Performances of online linear regression IFLH algorithms and OGD with fixed restart on the time series generated with the Soft Shifts for various dimension $d$ and parameter $\alpha$}
    \label{fig:lr-ss12d_all}
\end{figure*}

\begin{figure*}[t]
 \centering
    \begin{subfigure}[t]{0.32\textwidth}
        \centering
        \includegraphics[width=\textwidth]{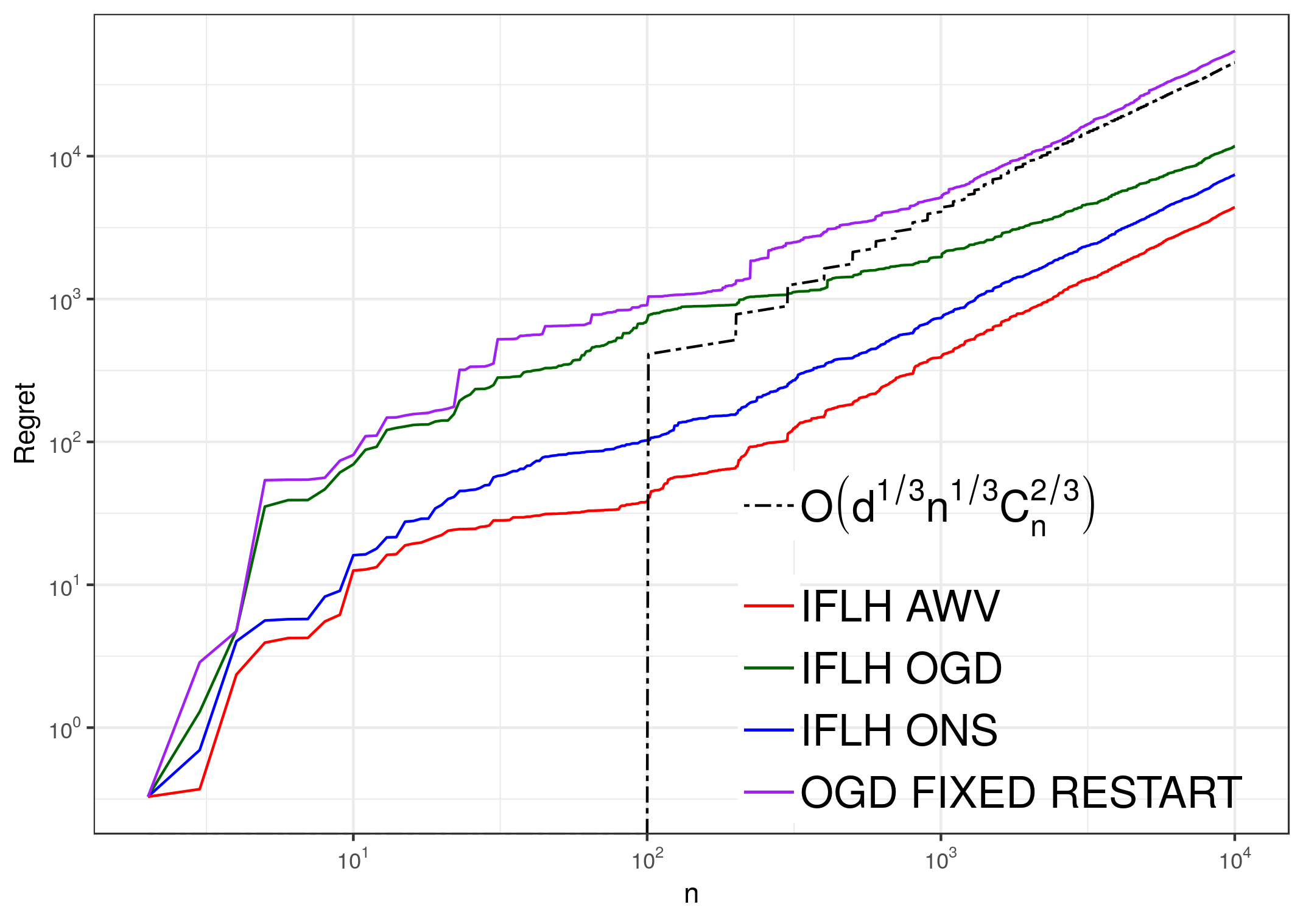}
        \caption{$\{m_i = 100i \,|\, 1 \leq i \leq 100\}$,  $d = 10$}
     \label{fig:lr-hsd10}
    \end{subfigure}~ 
    \begin{subfigure}[t]{0.32\textwidth}
        \centering
        \includegraphics[width=\textwidth]{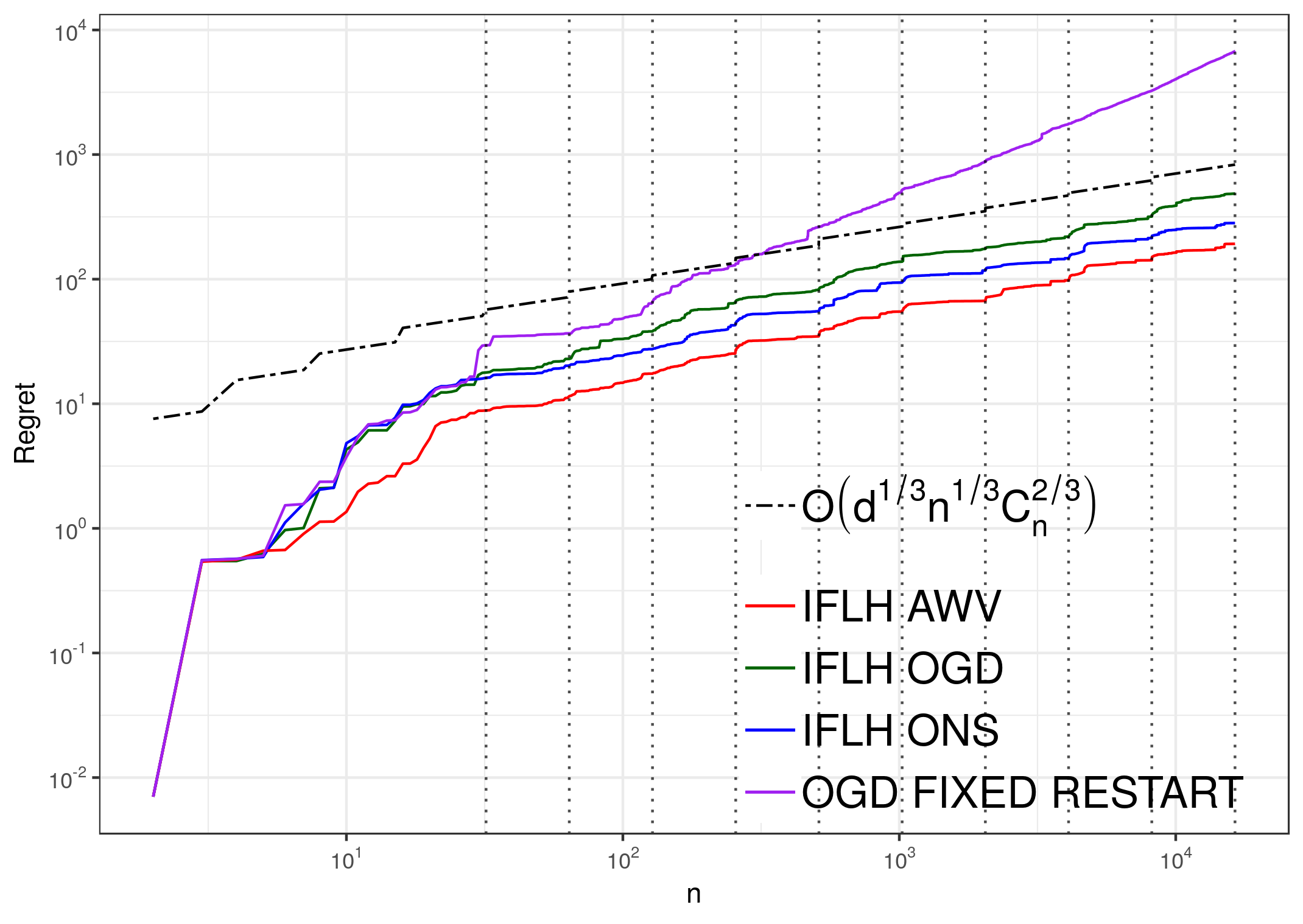}
        \caption{$\{m_i = 2^i \,|\, 1 \leq i \leq 14\}$,  $d = 2$}
     \label{fig:lr-hsd2log}
    \end{subfigure}~ 
    \begin{subfigure}[t]{0.32\textwidth}
        \centering
        \includegraphics[width=\textwidth]{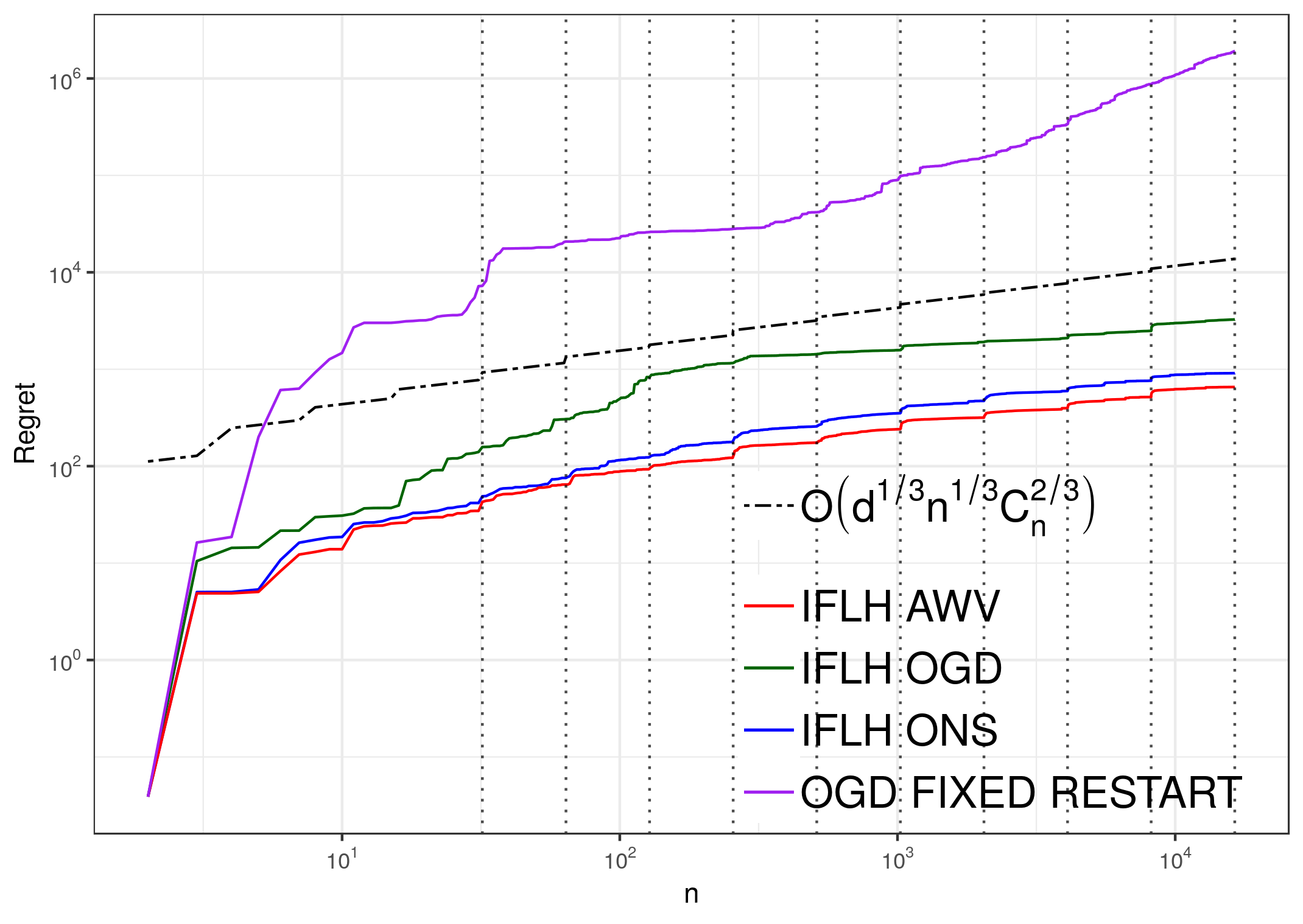}
        \caption{$\{m_i = 2^i \,|\, 1 \leq i \leq 14\}$,  $d = 10$}
     \label{fig:lr-hsd10log}
    \end{subfigure}
    \vspace{-2mm}
    \caption{Performances of online linear regression IFLH algorithms and OGD with fixed restart on the time series generated with the Hard Shifts.}
    \label{fig:lr-hsd-all}
\end{figure*}
\subsection{1-Dimension (Figs~\ref{figs:1d-pred} and \ref{figs:1d-regret})}
We use  ARROWS \citep{baby2019online} as our baseline for this part of the experiment, we compare it with our procedure proposed in Section~\ref{sec:warmup}, that is IFLH with moving averages as a subroutine.  We recall that ARROWS was especially designed for this one dimensional setting in which it achieves the optimal rate. It also requires the variance of the noise $\sigma^2$ to be give beforehand which is not the case for our procedure.
We average the predictions and the cumulative errors on 10 iterations over the time series. In all of our experiments, we consider the sub-Gaussian noise with standard deviation to be $\sigma = 1$. We have $y_t = \theta_t + Z_t$ with $Z_t \sim \mathcal{N}(0, \sigma^2)$. We generate data by soft shifting and hard shifts mechanism described above.  
\paragraph{Soft Shifts:} In first part of our experiment, we generate the data by soft shifting mechanism. The parameter $\alpha$, which controls how much the time-series is non-stationary, is set to be $0.3$. This results in a slow decay of total variation of order $\sum_{t=2}^n t^{-\alpha} = O(n^{1-\alpha}) = O(n^{0.7})$ and in an upper-bound of order $O(C_n^{2/3}n^{1/3}) = O(n^{1-\frac{2}{3}\alpha}) = O(n^{4/5})$. We can see in Figure~\ref{figs:1d-regret-ss} that IFLH reacts faster to slight changes in the time series yielding a slightly smaller cumulative error than ARROWS.  
\paragraph{Hard Shifts:}For the second part of the experiment, we generate data using the hard shift mechanism. In Figures \ref{figs:1d-pred-hs} and \ref{figs:1d-regret-hs} we test IFLH and ARROWS on a time series with equal spaced shifts, whereas in Figure \ref{figs:1d-pred-hslog} and \ref{figs:1d-regret-hslog}, time intervals between shifts grow exponentially with the length of the total number of shifts. It is clear from the plots that IFLH reacts faster than ARROWS to abrupt changes and manages to adapt better to stationary portions of the time series.
\subsection{Online linear regression}
We test IFLH \cite{zhang2017dynamic} on the online linear regression setting with three different subroutines: Online Gradient Descent (OGD), Online Newton Steps (ONS) as well as AWV (online ridge regression). We chose these subroutines because they are well used by the online learning community for standard stationary online linear regression. Note that ONS and AWV achieve optimal regret while this is not the case for OGD which cannot take advantage of the exp-concavity of the square loss. We compare their performances with the Online Gradient Descent with fixed restart of \citet{besbes2015non}. We use as batch size their theoretical result of $\left \lceil{\sigma^{-1}\sqrt{n\log n} / TV} \right \rceil $. We again consider two data generation mechanism as described above (soft shifting and hard shifting) to generate decision vectors $\theta_t$ for all~$t$. We take the sub-Gaussian noise to be multivariate normal with $\Sigma = I_d$. We have
\[
  Y_t = X_t^\top \theta_t + Z_t 
\]
with $Z_t \sim \mathcal{N}(0, \Sigma)$. We take $X_t$ to be multivariate uniformly distributed random variables $X_t \sim U( -\mathbf{1}, \mathbf{1})$. The expected cumulative error of OGD with fixed restart grows at a rate greater than the the theoretical upper bound of $O(d^{1/3}n^{1/3}TV^{2/3})$ proved in this paper. IFLH algorithms regrets stay below the theoretical upper bound.
\paragraph{Soft shifts:} In Figure~\ref{fig:lr-ss12d_all}, we vary the noise decaying parameter $\alpha$ as well as the dimension $d$ of the time series. We can clearly notice the better performance of IFLH algorithms especially with ONS and AWV as subroutine. When $\alpha = 2$ for instance, the sequence of $\theta_t$ quickly converges and OGD with fixed restart continues on resetting which leads to the high divergence of its regret. 
\paragraph{Hard shifts:}In experiment \ref{fig:lr-hsd10}, we use fixed size chunks. The OGD with fixed restart algorithm performs well since the sizes of the chunks are constant and adopting a fixed restart window strategy corresponds to the setting. IFLH algorithms reacts faster to these changes and have a slightly lower regret. In \ref{fig:lr-hsd2log} and \ref{fig:lr-hsd10log}, we use an exponentially growing size partitions. OGD with fixed restart's regret grows at a rate bigger than the boundary line of $O(d^{1/3}n^{1/3}TV^{2/3})$. IFLH algorithms conserve a regret rate of this order.

\paragraph{Acknowledgments} This work was funded in part by the French government under management of Agence Nationale de la Recherche as part of the "Investissements d’avenir" program, reference ANR-19-P3IA-0001 (PRAIRIE 3IA Institute).

\newpage
\small
\bibliography{online-ml.bib}
\bibliographystyle{plainnat}
\normalsize

\newpage
\onecolumn
\appendix
\begin{center}
{\centering \LARGE Appendix }
\vspace{1cm}
\sloppy

\end{center}

\section{Warmup : One Dimensional Time Series}\label{app:adaptive_restart_1d}

\label{app:proof_approx}

\begin{theorem}[Approximation error]
\label{thm:approx_error}
Let $n, m \geq 1$, $\sigma >0$, and $C_n >0$. Assume that $1\leq t_1,\dots,t_{m+1} = n+1$ are defined such that~\eqref{eq:fixed_batches} holds for each $1\leq i \leq m$. Then, for any sequence $\theta_1,\dots,\theta_n$ such that $TV(\theta_{1:n}) \leq C_n$ and $|\theta_1|\leq B$, the hypothetical forecasts $\tilde y_t$ defined in  Equation~\eqref{eq:constant_pred} satisfy
\[
  R_n(\tilde y_{1:n}, \theta_{1:n}) \leq  B^2 + TV(\theta_{1:n})^2  + 2 m \sigma^2 (2+\log n)    + \frac{n}{m^2} TV(\theta_{1:n})^2\,.
\]
Therefore, optimizing  $m \eqdef \left( \frac{n C_n^2}{\sigma^2(2+\log n)} \right)^{1/3}$ yields\footnote{Throughout the paper, the notation $\lesssim$ denotes a rough inequality which is up to universal multiplicative or additive constants and poly-logarithmic factors in $n$.}
\[
  R_n(\tilde y_{1:n}, \theta_{1:n}) \lesssim B^2 + C_n^2 +  n^{1/3} C_n^{2/3}\sigma^{4/3} (2+\log n)^{2/3} \,.
\]
\end{theorem}

\begin{proof}

Let $\tilde y_t$ be the estimate of the restarted moving average forecaster defined in Eq.~\eqref{eq:constant_pred}  at time $t$. Let $m \geq 1$ be the total number of batches and $1= t_1 \leq \dots\leq t_{m+1} = n+1$  and batches be numbered as $1,\cdots, m$ where $m$ is the total number of batches. By Equation~\eqref{eq:fixed_batches}, the total variation of ground truth within batch $i$ is fixed and is bounded by $\frac{C_n+B}{m}$ for each $i$, \textit{i.e.} if the time interval of batch $i$ is denoted by $[t_i, t_{i+1}-1]$ then by Inequality~\eqref{eq:fixed_batches}
\[
    \sum_{t = t_i}^{t_{i+1}-2} | \theta_{t} - \theta_{t+1}| \leq \frac{C_n}{m} \,.
\]
Let us fix a batch $i \in \{1,\dots,m\}$. By~\eqref{eq:constant_pred}, the cumulative error within the  batch equals
\begin{align*}
    R_i & := \sum_{t = t_i}^{t_{i+1}-1} \bbE \left[(\tilde y_t - \theta_t)^2 \right] \\
    & \stackrel{\eqref{eq:constant_pred}}{=} \bbE \left[(\bar y_{t_{i-1}:(t_i-1)} - \theta_{t_i})^2 \right] + \sum_{t=t_i+1}^{t_{i+1}-1}  \bbE \left[\Big( \bar y_{t_i:(t-1)} - \theta_t \Big)^2 \right] \\
    & = \bbE \left[(\bar \theta_{t_{i-1}:(t_i-1)} + \bar Z_{t_{i-1}:(t_i-1)} - \theta_{t_i})^2 \right] + \sum_{t=t_i+1}^{t_{i+1}-1}  \bbE \left[\Big( \bar \theta_{t_i:(t-1)} + \bar Z_{t_i:(t-1)} - \theta_t \Big)^2 \right] 
\end{align*}
where the notation $\bar x_{t:t'}$ means $\sum_{s=t}^{t'} x_s$ and where we used that $y_t = \theta_t + Z_t$ for any $t$. Using that $Z_t$ are i.i.d. random variables with $\E[Z_t] = 0$ and $\E[Z_t^2] \leq \sigma^2$, we have by bias-variance decomposition
\begin{align*}
    R_i & =  (\bar \theta_{t_{i-1}:(t_i-1)}  - \theta_{t_i})^2  + \bbE\big[\bar Z_{t_{i-1}:(t_i-1)}^2\big] + \sum_{t=t_i+1}^{t_{i+1}-1}  \big( \bar \theta_{t_i:(t-1)}  - \theta_t \big)^2  +  \E\big[\bar Z_{t_i:(t-1)}^2\big]  \\
    & \leq (\bar \theta_{t_{i-1}:(t_i-1)}  - \theta_{t_i})^2  + \frac{\sigma^2}{t_{i} - t_{i-1}} + \sum_{t=t_i+1}^{t_{i+1}-1}  \big( \bar \theta_{t_i:(t-1)}  - \theta_t \big)^2  +  \frac{\sigma^2}{t - t_i}
\end{align*}
Assuming $\theta_0 = 0$, and summing across all bins yields that the cumulative error is upper-bounded by,
\begin{align*}
     R_n(\tilde y_{1:n}, \theta_{1:n}) := \sum_{i=1}^m R_i 
        & \leq \sum_{i=1}^m (\bar \theta_{t_{i-1}:(t_i-1)}  - \theta_{t_i})^2  + \sum_{i=1}^m \sum_{t=t_i+1}^{t_{i+1}-1}  \Big( \bar \theta_{t_i:(t-1)}  - \theta_t \Big)^2  + \sum_{i=1}^m \sum_{t=t_i+1}^{t_{i+1}} \frac{\sigma^2}{t - t_i} \\
        & \stackrel{t_1 = 1}{\leq}   |\theta_1|^2 + \bigg(\sum_{i=2}^m \big|\bar \theta_{t_{i-1}:(t_i-1)}  - \theta_{t_i}\big| \bigg)^2  + \sum_{i=1}^m   \sum_{t=t_i+1}^{t_{i+1}-1}  \big(\bar \theta_{t_i:(t-1)}  - \theta_t\big)^2  + \sum_{i=1}^m \sum_{t=t_i+1}^{t_{i+1}} \frac{\sigma^2}{t - t_i} \\
        & \leq  B^2 + \bigg(\sum_{i=2}^m \big|\bar \theta_{t_{i-1}:(t_i-1)}  - \theta_{t_i}\big| \bigg)^2  + \sum_{i=1}^m   \sum_{t=t_i+1}^{t_{i+1}-1}  \big(\bar \theta_{t_i:(t-1)}  - \theta_t\big)^2  + 2 m \sigma^2 (2+\log n)
\end{align*}
Then, because for all $i\geq 1$ and $t_{i+1} \geq t \geq t_i$,
\begin{multline*}
    \big| \bar \theta_{t_i:(t-1)}  - \theta_t \big|   
         = \bigg| \frac{1}{t-t_i} \sum_{s=t_i}^{t-1} \theta_s - \theta_t\bigg|
        \stackrel{\text{Jensen}}{\leq} \frac{1}{t-t_i} \sum_{s=t_i}^{t-1} \big| \theta_s - \theta_t\big| \leq \max_{s \in \{t_i,\dots,t-1\}} |\theta_s - \theta_t| \\
         =  \max_{s \in \{t_i,\dots,t-1\}}  \bigg| \sum_{r = s}^{t-1} \theta_{r} - \theta_{r+1} \bigg|
        \leq \max_{s \in \{t_i,\dots,t-1\}}   \sum_{r = s}^{t-1} \big| \theta_{r} - \theta_{r+1} \big|
        = \sum_{s=t_i}^{t-1} |\theta_s - \theta_{s+1}|\,,
\end{multline*}
we have
\begin{align*}
     R_n(\tilde y_{1:n}, \theta_{1:n}) 
        & \leq B^2 + \bigg(\sum_{i=2}^m  \sum_{s=t_{i-1}}^{t_i-1} |\theta_s - \theta_{s+1}| \bigg)^2  + \sum_{i=1}^m  \sum_{t=t_i+1}^{t_{i+1}-1}  \bigg( \sum_{s=t_i}^{t-1} |\theta_s - \theta_{s+1}| \bigg)^2   + 2 m \sigma^2 (2+\log n) \\ 
        & \leq B^2 + C_n^2  + \sum_{i=1}^m  \sum_{t=t_i+1}^{t_{i+1}-1}  \bigg( \sum_{s=t_i}^{t-1} |\theta_s - \theta_{s+1}| \bigg)^2  + 2 m \sigma^2 (2+\log n) \,.
\end{align*}
Therefore, using Inequality~\eqref{eq:fixed_batches},
\begin{align*}
     R_n(\tilde y_{1:n}, \theta_{1:n}) 
        & \leq B^2 + C_n^2  + \sum_{i=1}^m  \sum_{t=t_i+1}^{t_{i+1}-1}  \bigg(\frac{C_n}{m} \bigg)^2  + 2 m \sigma^2 (2+\log n)  \\
    &\leq   B^2 + C_n^2  + \frac{C_n^2}{m^2} \sum_{i = 1}^m \big(t_{i+1} - t_i\big) + 2 m \sigma^2 (2+\log n) \\
     &\leq   B^2 + C_n^2  + \frac{ n C_n^2}{m^2} + 2 m \sigma^2 (2+\log n).
\end{align*}

Now in the above equation, the choice $m = \left( \frac{n C_n^2}{\sigma^2(2+\log n)} \right)^{1/3}$ yields
\begin{align}
     R_n(\tilde y_{1:n}, \theta_{1:n}) \leq B^2 + C_n^2 + 2 n^{1/3} C_n^{2/3}\sigma^{4/3} (2+\log n)^{2/3}.
\end{align}

\end{proof}
\begin{remark}
Since, we also have the boundedness assumption here on each $\theta_i$ such that $|\theta_i| \leq B$ hence, it is easy to see that the bound given in the above result in Theorem~\ref{thm:approx_error} can be written as
\begin{align}
    R_n(\tilde y_{1:n}, \theta_{1:n}) \leq B^2 + 2BC_n + 2 n^{1/3} C_n^{2/3}\sigma^{4/3} (2+\log n)^{2/3}. 
\end{align}
\end{remark}




\section{Non-Stationary Online Linear  Regression} \label{ap:linear}

\subsection{Bias-variance decomposition for online linear regression}

\begin{lemma}[Restatement of Lemma~\ref{lem:bias_variance_square}]
For any sequence of functions $\tilde{g}_t : \mathbb{R}^d\rightarrow \mathbb{R}$ for $t \in [n]$ independent of $Z_t$ for all $t$, the cumulative error $R_n(\hat y_{1:n}, g_{1:n}) $ can be decomposed as follows:
\begin{align*}
R_n(\hat y_{1:n},g_{1:n})= \sum_{t=1}^n \bbE \left[(\hat{y}_t - {y}_t)^2 - (\tilde{g}_t(x_t) - {y}_t)^2 \right] + \sum_{t=1}^n \bbE\left[ (\tilde{g}_t(x_t) - {g}_t(x_t))^2\right].
\end{align*}
\end{lemma}
\begin{proof}
Let $t\geq 1$. Since $y-t - g_t(x_t) = Z_t$, which is zero mean and independent from $\hat g_t(x_t) - g_t(x_t)$, we have
\begin{equation}
    \bbE\Big[\big(\hat g_t(x_t) - y_t\big)^2\Big] =  \bbE\Big[\big(\hat g_t(x_t) - g_t(x_t) + g_t(x_t) -  y_t\big)^2\Big] = \bbE\Big[\big(\hat g_t(x_t) - g_t(x_t)\big)^2\Big] + \bbE\Big[ \big(g_t(x_t) -  y_t\big)^2\Big].
    \label{eq:decomp_bias_variance}
\end{equation}
Therefore, by definition~\eqref{eq:gen_non-station} of the cumulative error
\begin{eqnarray*}
R_n(\hat y_{1:n},g_{1:n}) &\stackrel{\eqref{eq:gen_non-station}}{=}& \sum_{t=1}^n \bbE \left[(\hat{g}_t(x_t) - {g}_t(x_t))^2\right]  \\
&\stackrel{\eqref{eq:decomp_bias_variance}}{=}& \sum_{t=1}^n \bbE \left[(\hat{g}_t(x_t) - {y}_t)^2  - ({g}_t(x_t) - {y}_t)^2\right] \notag \\
&=& \sum_{t=1}^n \bbE \left[(\hat{g}_t(x_t) - {y}_t)^2 - (\tilde{g}_t(x_t) - {y}_t)^2 + (\tilde{g}_t(x_t) - {y}_t)^2  - ({g}_t(x_t) - {y}_t)^2\right]  \notag \\
&= &\sum_{t=1}^n \bbE \left[(\hat{g}_t(x_t) - {y}_t)^2 - (\tilde{g}_t(x_t) - {y}_t)^2 \right] +\sum_{t=1}^n \bbE \left[  (\tilde{g}_t(x_t) - {y}_t)^2  - ({g}_t(x_t) - {y}_t)^2\right] \notag \\
&=& \sum_{t=1}^n \bbE \left[(\hat{g}_t(x_t) - {y}_t)^2 - (\tilde{g}_t(x_t) - {y}_t)^2 \right] +\sum_{t=1}^n \bbE \left[  \tilde{g}_t(x_t)^2 - 2\tilde{g}_t(x_t) {y}_t   - {g}_t(x_t)^2 +2 {g}_t(x_t) {y}_t\right] \notag \\
&\stackrel{\eqref{eq:generalized_curve_fitting}}{=}& \sum_{t=1}^n \bbE \left[(\hat{g}_t(x_t) - {y}_t)^2 - (\tilde{g}_t(x_t) - {y}_t)^2 \right] +\sum_{t=1}^n \left[\bbE [  \tilde{g}_t(x_t)^2] - 2\bbE[\tilde{g}_t(x_t) (g_t(x_t) + Z_t )] \right. \notag \\
&& \qquad \qquad \qquad \qquad \qquad \qquad \qquad \qquad  \qquad  \qquad  \left. - \bbE[{g}_t(x_t)^2] +  \bbE[2{g}_t(x_t)(g_t(x_t) + Z_t)]  \right]  \notag \\
&=& \sum_{t=1}^n \bbE \left[(\hat{g}_t(x_t) - {y}_t)^2 - (\tilde{g}_t(x_t) - {y}_t)^2 \right] + \sum_{t=1}^n \bbE\left[ (\tilde{g}_t(x_t) - {g}_t(x_t))^2\right] \,,
\end{eqnarray*}
where the last line of the proof comes from the fact that $\tilde{g}_t(x_t)$ is independent of $Z_t$ for all $t$.
\end{proof}

\subsection{Approximation error of the hypothetical forecaster}

\begin{lemma}[Restatement of Lemma~\ref{lem:approx_least_square}]
\label{lem:approx_least_square_bis}
Let $X,B>0$. Assume that $\|x_t\|\leq X$ and $\|\theta_t\| \leq B$ for all $t\in [n]$. Then, there exists a sequence of restarts $1= t_1<\dots<t_m = n+1$ such that
\[
   \sum_{t=1}^n (x_t^\top \bar \theta_t - x_t^\top  \theta_t )^2  =  \sum_{j=1}^m \sum_{t=t_j}^{t_{j+1}-1} \big((\bar \theta_{t_j:(t_{j+1}-1)}  - \theta_t )^\top x_t\big)^2   \leq X^2 n \Big(\frac{C_n}{m}\Big)^2 + 4 X^2 B^2 m \,,
\]
where 
\[
    \bar \theta_t := \bar \theta_{t_j:(t_{j+1}-1)} \quad \text{for} \quad  t_j \leq t \leq t_{j+1} - 1 \qquad \text{and} \qquad  \bar \theta_{t_j:(t_{j+1}-1)}  = \frac{1}{t_{j+1} -t_j} \sum_{t=t_j}^{t_{j+1}-1} \theta_t \,.
\]
\end{lemma}

\begin{proof}

Let $m \in [n]$ be the total number of batches. Let $1=t_1 \leq \dots \leq t_{m+1} = n+1$ be such that the total variation of the ground truth with each batch $i$ is at most $(C_n+B)/m$, that is for all $i \in [m]$
\begin{equation}
    \label{eq:batch_linear_reg}
    \sum_{t=t_i}^{t_{i+1}-2} \big\|\theta_t - \theta_{t+1}\big\|_1 \leq \frac{C_n}{m} \,.
\end{equation}
Therefore, 
\begin{align*}
 \sum_{t=t_i}^{t_{i+1} -1} \E\big[ (x_t^\top  \bar \theta_{t_i:(t_{i+1}-1)}  -  x_t^\top \theta_t)^2\big] 
    & \leq \sum_{t=t_i}^{t_{i+1} -1} \E\big[\| \bar \theta_{t_i:(t_{i+1}-1)}  - \theta_t\|_2^2 \|x_t\|^2\big] \\
    & \leq X^2 \sum_{t=t_i}^{t_{i+1} -1} \| \bar \theta_{t_i:(t_{i+1}-1)}  - \theta_t\|_2^2  \\
    & \leq 4 X^2 B^2 + X^2 \sum_{t=t_i}^{t_{i+1} -2} \| \bar \theta_{t_i:(t_{i+1}-1)}  - \theta_t\|_2^2 \\
    & \leq 4 X^2 B^2 + X^2 \sum_{t=t_i}^{t_{i+1} -2} \big\| \bar \theta_{t_i:(t_{i+1}-1)}  - \theta_t\big\|_1^2 \,.
\end{align*}
But, since for all $i \geq 1$ and all $t \in \{t_i,\dots,t_{i+1}-2\}$
\[
      \big\| \bar \theta_{t_i:(t_{i+1}-1)}  - \theta_t\big\|_1 \stackrel{\text{Jensen}}{\leq} \frac{1}{t_{i+1}-t_i} \sum_{s=t_i}^{t_{i+1}-1} \big\|\theta_s - \theta_t\big\|_1 \leq \max_{t_i \leq s\leq t_{i+1}-1}  \big\|\theta_s - \theta_t\big\|_1  \leq \sum_{t=t_i}^{t_{i+1}-2} \big\|\theta_t - \theta_{t+1} \big\|_1 \stackrel{\eqref{eq:batch_linear_reg}}{\leq} \frac{C_n}{m} \,,
\]
it yields
\begin{equation}
  \sum_{t=t_i}^{t_{i+1} -1} \E\big[ (x_t^\top  \bar \theta_{t_i:(t_{i+1}-1)}  -  x_t^\top \theta_t)^2\big]   \leq  4X^2B^2 + X^2 (t_{i+1}-t_i-1) \Big(\frac{C_n}{m}\Big)^2.
  \label{eq:approx_error_batch_lin}
\end{equation}
Summing over all batches $i =1,\dots,m$ concludes the proof. 
\end{proof}

\subsection{Dynamic regret bound for IFLH with Online Newton Step}
We present here a result from \citet{zhang2017dynamic} on the adaptive regret of Algorithm~\ref{alg:iflh_d_1} that will be usefull for our regret analysis. Let us first recall their setting on non stationary online convex optimization. Let $\Omega \subset \mathbb{R}^d$ be a convex compact subset of $\mathbb{R}^d$. A sequence of convex loss functions $f_t:\Omega \to \mathbb{R}^d$ is sequentially optimized as follows. At each round $t=1,\dots, n$, a learner chooses a parameter $\theta_t \in \Omega$, then observes a subgradient $\nabla f_t(\theta_t)$ and updates $\theta_{t+1}$.  Learner's goal is to minimize his adaptive regret defined as the maximum static regret over intervals of length $\tau \geq 1$
\[
    \text{SA-Regret}(n,\tau) := \max_{1\leq s\leq n-\tau} \left\{ \sum_{t=s}^{s+\tau - 1} f_t(\theta_t) - \min_{\theta \in \Omega} \sum_{t=s}^{s+\tau -1} f_t(\theta) \right\} \,.
\]


\begin{theorem}[Theorem 1, \citep{zhang2017dynamic}] \label{thm:zhang_thm1}
Let $n, d \geq 1$, $\Omega \subseteq \mathbb{R}^d$, and $G, B, \alpha >0$. Let $f_1,\dots, f_n:\Omega \to \mathbb{R}^d$ be a sequence of $\alpha$-exp-concave  loss functions such that $\|\nabla f_t(\theta)\| \leq G$ for all $\theta \in \Omega$ and $1\leq t\leq n$.  Then, Algorithm~\ref{alg:iflh_d_1} (i.e., Alg.~1 of \citet{zhang2017dynamic} with $K=2$) with $\eta = \alpha$ and Online Newton Step as subroutine satisfies
\begin{align*}
    \text{SA-Regret}(T,\tau) \leq \left( \frac{(5d+1)( \lceil \log_2 \tau \rceil +1) +2}{\alpha}+5d(\lceil \log_2 \tau \rceil +1) GB\right) \log n =\mathcal{O}\left(d\log^2 n\right) \,,
\end{align*}
for any $\tau \in [n]$. 

\end{theorem}

\subsection{Proof of Theorem~\ref{thm:linear_non-station_reg}}

\begin{theorem}[Restatement of Theorem~\ref{thm:linear_non-station_reg}]
Let $n, m \geq 1$, $\sigma >0$, $B>0$, $X>0$, and $C_n >0$. Let $\theta_1,\dots,\theta_n$ such that $TV(\theta_{1:n}) \leq C_n$ and $\|\theta_t\|\leq B$. Assume that $\|x_t\|\leq X$ for all $t\geq 1$. Then, Alg.~\ref{alg:iflh_d_1} \cite{zhang2017dynamic} with Online Newton Step \cite{hazan2007logarithmic} as subroutine and well-tuned learning rate $\eta>0$  satisfies
\[
  R_n(\hat y_{1:n}, \theta_{1:n}) \lesssim  d^{1/3}n^{1/3} C_n^{2/3} (X^2\sigma^2 B + X^2 B^2)^{1/3}  \,,
\]
with high probability. 
\end{theorem}
\begin{proof}
As discussed before, here the goal is to control the expected cumulative error with respect to the unobserved outputs $\tilde y_t = x_t^\top \theta_t = y_t - Z_t$. Our prediction for $\theta_t$ at any time instant $t$ is denoted as $\hat{\theta}_t$. Hence, the prediction for $\tilde y_t$ is given by $\hat{y}_t = \hat \theta_t^\top  x_t $ and the expected cumulative error $R_n(\hat y_{1:n}, \theta_{1:n}) $ can be written as 
\[
    R_n(\hat y_{1:n}, \theta_{1:n}) = \sum_{t=1}^n \E\Big[(\hat y_t - \tilde y_t)^2 \Big] = \sum_{t=1}^n \E\Big[\big((\hat \theta_t - \theta_t)^\top x_t\big)^2 \Big] \,.
\]
Let $1=t_1\leq \dots\leq t_{m+1} = n+1$, $\bar \theta_t$, and $\bar \theta_{t_i:(t_{i+1}-1)}$ be defined as in Lemma~\ref{lem:approx_least_square_bis}. Applying Lemma~\ref{lem:bias_variance_square} with $\tilde{g}_t(x_t) = \bar{\theta}_t^\top x_t $ for all $t$, we have
\begin{eqnarray}
 R_n(\hat y_{1:n}, \theta_{1:n}) 
    &  = & \sum_{t =1}^n \E\left[ (x_t^\top \hat \theta_t - y_t)^2 - (x_t^\top \bar \theta_t - y_t)^2  \right]  + \sum_{t=1}^n  \Big[ (x_t^\top \bar \theta_t - x_t^\top  \theta_t)^2   \Big] \nonumber \\ 
 & \leq & \sum_{t =1}^n \E\left[ (x_t^\top \hat \theta_t - y_t)^2 - (x_t^\top \bar \theta_t - y_t)^2  \right]  + X^2 n \Big(\frac{C_n}{m}\Big)^2 + 4 X^2 B^2 m  \label{eq:cumerror} \,,
\end{eqnarray}
where the second inequality is by Lemma~\ref{lem:approx_least_square_bis}. Now, we can upper-bound the first term of the right-hand-side by applying  Theorem~\ref{thm:zhang_thm1} with $f_t(\theta) = (x_t^\top \theta - y_t)^2$. Then, 
\begin{align*}
\nabla f_t(\theta)  = 2(x_t^\top \theta - y_t) x_t  = 2(x_t^\top \theta - x_t^\top \theta_t - Z_t) x_t  = 2(x_tx_t^\top(\theta - \theta_t) ) - 2 Z_t x_t.
\end{align*}
Since for all $t\geq 1$, $Z_t$ are $\sigma$-subGaussian with zero-mean, we have
\[
    |Z_t| \leq 2 \sigma \sqrt{\log \frac{n}{\delta}}, \qquad \text{for all} \quad  t=1,\dots,n \,,
\]
with probability at least $1-\delta$. Hence, with probability at least $1-\delta$, for all $t\in [n]$ and all $\|\theta\|\leq B$
\[
    |y_t| = |\theta_t^\top x_t + Z_t\big| \leq BX + 2 \sigma \sqrt{\log \frac{n}{\delta}} \quad \text{and} \quad \big\|\nabla f_t(\theta)\big\| \leq 4 X^2 B^2 + 2\sigma X \sqrt{\log \frac{n}{\delta}} \,.
\]
We consider this favorable event until the end of the proof. In particular, this implies that $G = 4 X^2 B^2 + 2\sigma X \sqrt{\log \frac{n}{\delta}}$ and  that all losses $f_t$ are $\alpha$-exp-concave with any parameter $\alpha \leq   \big(16B^2X^2 + 2\sigma^2\log \frac{n}{\delta}\big)^{-1}$. Applying Theorem~\ref{thm:zhang_thm1}, for the choice $\eta = \alpha$ in Alg.~\ref{alg:iflh_d_1}, we thus get
\begin{multline*} 
\sum_{t =1}^n \E\left[ (x_t^\top \hat \theta_t - y_t) - (x_t^\top \bar \theta_t - y_t)^2  \right]  = \sum_{i=1}^m \sum_{t=t_i}^{t_{i+1}-1}  \E\left[ (x_t^\top \hat \theta_t - y_t) - (x_t^\top \bar \theta_{t_i:(t_{i+1}-1)} - y_t)^2  \right] \\
 \stackrel{\text{Thm.~\ref{thm:zhang_thm1}}}{\leq}  m \left( \frac{(5d+1)( \lceil \log_2 n \rceil +1) +2}{\alpha}+5d(\lceil \log_2 n \rceil +1) GB\right) \,.
\end{multline*}
Finally, substituting $G$ and $\alpha$, and plugging back into Inequality~\eqref{eq:cumerror}, we get
\begin{align*}
R_n(\hat y_{1:n}, \theta_{1:n}) \leq m \left( (5d+3)(16B^2X^2 + 2\sigma^2\log\frac{n}{\delta})+5d(4 X^2 B + 2 \sigma X\sqrt{ \log \frac{n}{\delta}})B\right) \log^2 n \\
+  X n \Big(\frac{C_n}{m}\Big)^2 + 4X^2B^2m \,,
\end{align*}
with probability greater than $1 - \delta$. Choosing $m = \tilde{\mathcal{O}} \left( \frac{n^{1/3}C_n^{2/3}}{d^{1/3}(X^2\sigma^2 B + X^2 B^2)^{1/3}} \right)$ we get, 
\begin{align*}
R_n(\hat y_{1:n}, \theta_{1:n})  \leq \tilde{\mathcal{O}}\left( d^{1/3}n^{1/3} C_n^{2/3} (X^2\sigma^2 B + X^2 B^2)^{1/3} \right) \,.
\end{align*}
with high probability. 
\end{proof}

\section{Non-Stationary Online Kernel  Regression} \label{ap:kernel}
Below, we provide two results from \citet{jezequel2019efficient} for online kernel regression with square loss. Kernel-AWV  \citet{jezequel2020efficient} computes the following estimator.
\begin{align}
\hat{\theta}_t  = \argmin_{\theta \in \mathcal{H}} \left\{ \sum_{s=1}^{t-1} (y_s - \theta^\top \phi(x_t) )^2 +  \lambda \|\theta \|^2 + (\phi(x_t)^\top\theta )^2 \right\}, \label{eq:PKAWV}
\end{align}
where $\phi:\mathbb{R}^d \rightarrow \mathcal{H}$ and $\mathcal{H}$ is RKHS corresponding to kernel $\mathcal{K}$.
\begin{theorem}[Proposition 1, \citep{jezequel2019efficient}]  \label{thm:jezequel1}
Let $\lambda, Y \geq 0$, $\mathcal{X} \subset \mathbb{R}^d$ and $\mathcal{Y} \subset [-Y,Y]$.  For any RKHS $\mathcal{H}$,  for $n \geq 1$, for  any arbitrary sequence of observations $ (x_1, y_1), \cdots ,(x_n, y_n) \in \mathcal{X}\times \mathcal{Y}$, the regret of Kernel-AWV (Equation~\eqref{eq:PKAWV}, \cite{jezequel2020efficient}) is upper-bounded for all  $\theta \in \mathcal{H}$ as
\begin{align*}
R_n(\theta) := \sum_{t=1}^n (\hat y_t - y_t)^2 - (\theta^\top \phi(x_t) - y_t)^2 \leq \lambda \| \theta\|_{\mathcal{H}}^2  + Y^2 \sum_{k=1}^n \log \left(  1+ \frac{\lambda_k(K_{nn})}{\lambda}  \right)
\end{align*}
where  $K_{nn}$ is defined as $(K_{nn})_{i,j} = \langle\phi(x_i), \phi(x_j)\rangle$ and $\lambda_k(K_{nn})$ denotes the $k$-th largest eigenvalue of $K_{nn}$.
\end{theorem}

\begin{theorem}[Proposition 2, \citep{jezequel2019efficient}] \label{thm:jezequel2}
For all $n\geq 1$, $\lambda > 0$ and all input sequences $x_1, \cdots , x_n \in \mathcal{X}$, 
\begin{align*}
\sum_{k=1}^n \log \left(  1+ \frac{\lambda_k(K_{nn})}{\lambda}  \right) \leq \log \left( e + \frac{en\kappa^2}{\lambda}\right)d_{eff}(\lambda).
\end{align*}
where $\kappa = \sup_{x\in \mathcal{X}}\mathcal{K}(x,x)$ and $d_{eff}(\lambda) := \text{Tr}(K_{nn}(K_{nn}+\lambda I_n)^{-1})$.
\end{theorem}

Before proceeding to the next result, we reiterate our definition of time dependent effective dimension
\begin{align}
d_{eff}(\lambda,s,r) = Tr(K_{s-r,s-r}(K_{s-r,s-r}+\lambda I)^{-1})
\end{align}
where by abuse of notation $K_{s-r,s-r} = \phi(x_i)^\top\phi(x_j)$ for $r\leq i\leq s$ and $r \leq j \leq s$. It is also important to note that for each fixed $r$, $d_{eff}(\lambda,s,r) $ is an increasing function of $s-r$, so that we assume that their exists an upper-bound such that for all $1\leq r\leq s \leq n$,
\[
    d_{eff}(\lambda,s,r) \leq d_{eff}(\lambda, s-r) \,, 
\]
which only depends on $s-r$.

\begin{theorem}[Restatement of Theorem~\ref{thm:zhang_rkhs}]
For online kernel regression with square loss if for all $t \in [n]$, $y_t \in [-Y,Y]$, then for the Alg.~\ref{alg:iflh_d_1} with Kernel-AWV  \cite{jezequel2020efficient} as subroutine with parameter $\lambda>0$,  we have for all $1\leq r\leq s\leq n$ and all $\theta \in \mathcal{H}$
\begin{align*}
    \sum_{t=r}^s f_t(\theta_t) - \sum_{t=r}^s f_t(\theta) \leq 8Y^2({p+2})\log n +  \lambda p \| \theta\|^2 + Y^2p d_{eff}(\lambda,s-r) \log \left( e + \frac{en\kappa^2}{\lambda}\right) \,,
\end{align*}
where $p \leq \lceil \log_2(s-r+1) \rceil + 1$ and $f_t(\theta) = (y_t - \phi(x_t)^\top \theta)^2$.
\end{theorem}
\begin{proof}
Following the proof of Theorem~1 from \cite{zhang2017dynamic}, we know that there exists $p$ segments
\begin{align*}
I_j = [t_j, \tau_{t_j}],~ j \in [p]
\end{align*}
with $p \leq \lceil \log_{2} (s-r+1)\rceil +1 $, such that 
$t_1 = r$, $t_{j+1} = \tau_{t_j} + 1$, $j \in [p-1]$ and $\tau_{t_p} \geq s$. Also, the expert (or subroutine) $\mathcal{A}_{t_j}$ corresponds to Kernel-AWV started at round $t_j$ and stopped at round $\tau_{t_j}$. We denote  ${\theta}_{t_j}^{t_j}, \cdots, {\theta}_{\tau_{t_j}}^{t_j}$ as the sequence  of solutions generated by the subroutine $\cA_{t_j}$. In other words, $\theta_{t}^{t_j}$ denotes the prediction at round $t$ output by an instance of Kernel-AWV started at time $t_j$. Following the proof of Theorem~1 of \cite{zhang2017dynamic}, we have
\begin{align}
\sum_{j=1}^{p-1} \left( \sum_{t=t_j}^{\tau_{t_j}} f_t(\hat{\theta}_t) - f_t({\theta}_{t}^{t_j})  \right) + \sum_{t=t_p}^{s} f_t(\hat{\theta}_t) - f_t({\theta}_{t}^{t_p})  \leq \frac{1}{\alpha} \sum_{j=1}^p \log t_j + \frac{2}{\alpha} \sum_{t = r+1}^s \frac{1}{t} \leq \frac{p+2}{\alpha}\log n,
\end{align}
where $\alpha$ is the exp-concavity parameter of the functions $f_t$ that will be fixed later. 
From Theorem~\ref{thm:jezequel1} and \ref{thm:jezequel2}, for any $j \in [p-1]$, the regret of the subroutine $\cA_{t_j}$ can be upper-bounded as
\begin{align*}
\sum_{t=t_j}^{\tau_{t_j}} f_t(\theta_t^{t_j}) -  f_t(\theta) 
    & \leq \lambda \| \theta\|^2  + Y^2d_{eff}(\lambda,t_j, \tau_{t_j}) \log \left( e + \frac{en\kappa^2}{\lambda}\right)\\
    & \leq \lambda \| \theta\|^2  + Y^2d_{eff}(\lambda,\tau_{t_j}- t_j) \log \left( e + \frac{en\kappa^2}{\lambda}\right) \\
    & \leq \lambda \| \theta\|^2  + Y^2d_{eff}(\lambda,s- r) \log \left( e + \frac{en\kappa^2}{\lambda}\right)\,.
\end{align*}
Similarly for $j = p$, we have
\begin{align*}
\sum_{t=t_p}^{s} f_t(\theta_t^{t_m}) -  f_t(\theta) \leq \lambda \| \theta\|^2  + Y^2d_{eff}(\lambda,s-r) \log \left( e + \frac{en\kappa^2}{\lambda}\right).
\end{align*} 
Combining everything together, we have
\begin{align*}
\sum_{t=r}^s f_t(\hat{\theta}_t) - f_t(\theta)  \leq  \frac{p+2}{\alpha}\log n +  \lambda p \| \theta\|^2 + Y^2p d_{eff}(\lambda,s-r) \log \left( e + \frac{en\kappa^2}{\lambda}\right) \,.
\end{align*}
For square loss with bounded output domain \textit{i.e.} $y_i \in[-Y,Y]$ for all $i \in [n]$, the square loss is $\alpha$-exp-concave with $\alpha  = 1/8B^2$. Hence, substituting the value 
\begin{align*}
\sum_{t=r}^s f_t(\hat{\theta}_t) - f_t(\theta)  \leq 8Y^2({p+2})\log n +  \lambda p \| \theta\|^2 + Y^2p d_{eff}(\lambda,s-r) \log \left( e + \frac{en\kappa^2}{\lambda}\right).
\end{align*}
\end{proof}

\begin{lemma}[Restatement of Lemma~\ref{lem:approx_least_square_rkhs}] 
\label{lem:approx_least_square_rkhs_bis} 
Let $B, \kappa >0$. Assume that $\|\phi(x_t) \|^2 \leq \kappa^2$, and $\|\theta_t\|_{\mathcal{H}} \leq B$ for all $t$. Then, there exists a sequence of restarts $1= t_1<\dots<t_m = n+1$ such that
\[
   \sum_{t=1}^n (\phi(x_t)^\top \bar \theta_t - \phi(x_t)^\top  \theta_t )^2  =  \sum_{j=1}^m \sum_{t=t_j}^{t_{j+1}-1} \big((\bar \theta_{t_j:(t_{j+1}-1)}  - \theta_t )^\top \phi(x_t)\big)^2   \leq \kappa^2 n \Big(\frac{C_n}{m}\Big)^2 + 4\kappa^2 B^2 m\,,
\]
where $\bar \theta_t := \bar \theta_{t_j:(t_{j+1}-1)}$ for $t_j \leq t < t_{j+1}$, $C_n \geq \sum_{t=2}^n \|\theta_t - \theta_{t-1} \|_{\mathcal{H}}$, and
\[
    \bar \theta_{t_j:(t_{j+1}-1)}  = \frac{1}{t_{j+1} -t_j} \sum_{t=t_j}^{t_{j+1}-1} \theta_t.
\]
\end{lemma}

\begin{proof}
Let $m$ be the total number of batches and $1=t_1\leq \dots\leq t_{m+1} = n+1$ such that for each batch $i \in [m]$ the total variation within the batch is upper-bounded as
\[
    \sum_{i=t_j}^{t_{j+1} -2}\|\theta_t - \theta_{t+1}\|_{\mathcal{H}} \leq \frac{C_n}{m} \,.
\]
Following the proof of Lemma~\ref{lem:approx_least_square_bis}, we get for all $i \in [m]$
\begin{align*}
 \sum_{t=t_i}^{t_{i+1} -1} \E\big[ (\phi(x_t)^\top  \bar \theta_{t_i:(t_{i+1}-1)}  -  \phi(x_t)^\top \theta_t)^2\big]  & \leq \sum_{t=t_i}^{t_{i+1} -1} \E\big[\| \bar \theta_{t_i:(t_{i+1}-1)}  - \theta_t\|_{\mathcal{H}}^2 \|\phi(x_t)\|_{\mathcal{H}}^2\big] \\
    & \leq \kappa^2 \sum_{t=t_i}^{t_{i+1} -1}\E\big[ \| \bar \theta_{t_i:(t_{i+1}-1)}  - \theta_t\|_{\mathcal{H}}^2\big]  \\
    & \leq  4\kappa^2B^2 + \kappa^{2} (t_{i+1}-t_i-1) \Big(\frac{C_n}{m}\Big)^2,
\end{align*}
where the last inequality is obtained similarly to~\eqref{eq:approx_error_batch_lin}. Summing over the batches $i=1,\dots,m$ concludes the proof.
\end{proof}

\begin{theorem}[Restatement of Theorem~\ref{thm:kernel_non_station_reg}]
Let $n, m \geq 1$, $\sigma >0$, $B>0$, $\kappa >0$, and $C_n >0$. Assume that 
\[
    d_{eff}(\lambda,r,s) \leq \Big(\frac{s-r}{\lambda}\Big)^\beta \,,
\]
for all $1\leq r\leq s\leq n$. Let $\theta_1,\dots,\theta_n$ such that $TV(\theta_{1:n}) \leq C_n$ and $\|\theta_t\|_{\mathcal{H}} \leq B$ for all $t\geq 1$. Assume also that $\|\phi(x_t)\|_{\mathcal{H}} \leq \kappa$ for $t\geq 1$.  Then , for well chosen $\eta>0$, Alg.~\ref{alg:iflh_d_1} with Kernel-AWV using $\lambda = ({n}/{m})^{\frac{\beta}{\beta +1}}$ and $m \eqdef \mathcal{O}\big( C_n^{\frac{2(\beta+1)}{2\beta+3}} n^{\frac{1}{2\beta+3}}\big)$ satisfies 
\begin{small}
\begin{align*}
  R_n(\hat y_{1:n}, \theta_{1:n}) \leq  \tilde{\mathcal{O}} \left( C_n^{\frac{2(\beta+1)}{2\beta+3}} n^{\frac{1}{2\beta+3}} \left( \sigma^2 \log \frac{1}{\delta}   + B^2\kappa^2  \right)
  +  (C_n+B)^{\frac{2}{2\beta+3}} n^{\frac{2\beta+1}{2\beta+3}} B^{\frac{4(\beta+1)}{2\beta+3}}\kappa^{\frac{2}{2\beta+3}}\right)  \,,
\end{align*}
\end{small}
 with probability at least $1-\delta$.
\end{theorem}
\begin{proof}
Recall that the cumulative error $R_n(\hat y_{1:n}, \theta_{1:n}) $ can be written as 
\[
    R_n(\hat y_{1:n}, \theta_{1:n}) = \sum_{t=1}^n \E\Big[(\hat y_t - y_t)^2 \Big] = \sum_{t=1}^n \E\Big[\big((\hat \theta_t - \theta_t)^\top \phi(x_t)\big)^2 \Big] \,.
\]
Let $m$ to be fixed later and let $\bar \theta_t$ and $1= t_1<\dots<t_m = n+1$, for $t \in \{t_j, \cdots, t_{j+1}\}$ be as defined in Lemma~\ref{lem:approx_least_square_rkhs}.
Applying Lemma~\ref{lem:bias_variance_square} with $\tilde{g}_t(x_t) = \bar{\theta}_t^\top \phi(x_t) $ for all $t$, followed by Lemma~\ref{lem:approx_least_square_rkhs_bis}, we get
\begin{align}
 R_n(\hat y_{1:n}, \theta_{1:n})   &= \sum_{t =1}^n \E\left[ (\phi(x_t)^\top \hat \theta_t - y_t)^2 - (\phi(x_t)^\top \bar \theta_t - y_t)^2  \right]  + \sum_{t =1}^n  \E\Big[ (\phi(x_t)^\top \bar \theta_t - \phi(x_t)^\top  \theta_t)^2   \Big] \nonumber\\
 & \leq \underbrace{\sum_{t =1}^n \E\left[ (\phi(x_t)^\top \hat \theta_t - y_t)^2  - (\phi(x_t)^\top \bar \theta_t - y_t)^2  \right] }_{:= T_1} + \kappa^2 n \Big(\frac{C_n}{m}\Big)^2 + 4\kappa^2 B^2 m. \label{eq:T1}
\end{align}
Now, we upper-bound $T_1$ the first term of the right-hand-side by applying Theorem~\ref{thm:zhang_rkhs}. We only need to compute the upper-bound $Y$ which will hold with high probability. Since for all $t\geq 1$, $Z_t$ are $\sigma$-subGaussian with zero-mean, we have
\[
    |Z_t| \leq 2 \sigma \sqrt{\log \frac{n}{\delta}}, \qquad \text{for all} \quad  t=1,\dots,n \,,
\]
with probability at least $1-\delta$. We consider this favorable high probability event until the end of the proof. 
 Hence, $|y_t|  = |\theta_t^\top \phi(x_t)  +Z_t| \leq B\kappa + 2\sigma \sqrt{ \log \frac{n}{\delta}} := Y$ for all $t\in [n]$. Therefore, Theorem~\ref{thm:zhang_rkhs} entails
\begin{align}
T_1 & := \sum_{t=1}^n  \E\left[ (\phi(x_t)^\top \hat \theta_t - y_t)^2 - (\phi(x_t)^\top \bar \theta_t - y_t)^2  \right] \notag  \\
& = \sum_{i=1}^m \sum_{t=t_i}^{t_{i+1}-1}  \E\left[ (\phi(x_t)^\top \hat \theta_t - y_t)^2 - (\phi(x_t)^\top \bar \theta_{t_i:(t_{i+1}-1)} - y_t)^2  \right] \label{eq:inter_t1_rkhs} \\ 
& \leq 8m Y^2(\log_2(n) + 4)\log n + m  \lambda (\log_2(n) + 2) B^2 + Y^2 (\log_2(n) + 2) \log \left( e + \frac{en\kappa^2}{\lambda}\right) \sum_{i=1}^m  d_{eff}(\lambda,t_{i+1}-t_i)  \nonumber
\end{align}
From the capacity condition, we know that there exists $\beta \in (0,1)$ such that for all $\lambda >0$ and $n\geq 1$
\[
    d_{eff}(\lambda, n) \leq \Big(\frac{n}{\lambda}\Big)^\beta \,.
\]
Hence, using $(\log_2(n)+4)\log n \leq 8 \log^2n$ for $n\geq 1$ and $\log_2(n)+2 \leq 5\log n$ for $n\geq 2$ (the error bound is true for $n=1$), we get
\begin{align}
T_1 &\leq 64 m Y^2 \log^2 n  + \lambda m B^2 \log n  + 5 Y^2\log n~\log \left( e + \frac{en\kappa^2}{\lambda}\right)   \sum_{j=1}^m \Big(\frac{t_{j+1}-t_{j}}{\lambda}\Big)^\beta  \notag   \\
&\leq 64 m Y^2 \log^2 n  + \lambda m B^2 \log n  + 5Y^2\log n~\log \left( e + \frac{en\kappa^2}{\lambda}\right)    m^{1-\beta}\Big(\frac{n}{\lambda}\Big)^\beta\,.
\end{align}
Last line comes from the Jensen's inequality. In the above equation, we choose $\lambda = \Big(\frac{n}{m}\Big)^{\frac{\beta}{\beta +1}}$ to get the following,
\begin{align}
T_1 &\lesssim  m Y^2 \log^2 n  + B^2 \log n ~m^{\frac{1}{\beta+1}} n^{\frac{\beta}{\beta+1}} \left( 1 + \log \left( e + {e\kappa^2 m^{\frac{\beta}{\beta+1}} n^{\frac{1}{\beta+1}}}\right)  \right)\,.
\end{align}
Plugging back into Inequality~\eqref{eq:T1}, it yields
\begin{align}
R_n(\hat y_{1:n}, \theta_{1:n}) &\lesssim   m Y^2 \log^2 n  + B^2 \log n ~m^{\frac{1}{\beta+1}} n^{\frac{\beta}{\beta+1}} \left( 1 + \log \left( e + {e\kappa^2 m^{\frac{\beta}{\beta+1}} n^{\frac{1}{\beta+1}}}\right)  \right) \notag \\
&\qquad \qquad \qquad \qquad \qquad \qquad \qquad \qquad \qquad \qquad   +  \kappa^2 n \Big(\frac{C_n}{m}\Big)^2 + \kappa^2 B^2 m.
\end{align}
Choosing $m = \mathcal{O}\Big( (C_n)^{\frac{2(\beta+1)}{2\beta+3}} n^{\frac{1}{2\beta+3}}\Big)$ concludes the proof.
\end{proof}

 \clearpage

\end{document}